\documentclass[twoside,11pt]{article}
\usepackage{jair, theapa, rawfonts}
\usepackage{amsmath}
\usepackage{amssymb}
\usepackage{amsfonts}
\usepackage{amssymb}
\usepackage{calligra}
\usepackage{calrsfs}

\let\emptyset\varnothing
\usepackage{relsize}
\usepackage[mathscr]{euscript}
\usepackage{blindtext} 
\usepackage{epstopdf}
\usepackage{booktabs}
\usepackage{listings}
\usepackage{tabularx}
\usepackage{subcaption}
\usepackage{graphicx}
\usepackage{comment}
\usepackage{todonotes}
\usepackage{multirow}
\usepackage{lscape}
\usepackage{subcaption} 
\usepackage{booktabs}
\usepackage{caption}

\usepackage{amsthm}
\newtheorem{theorem}{Theorem}[section]

\theoremstyle{definition} 
\newtheorem{definition}[theorem]{Definition}
\newtheorem{example}{Example}[section]

\usepackage{algorithm}
\usepackage{algorithmic}
\jairheading{1}{2024}{1-59}{09/2024}{}
\ShortHeadings{Towards Explainable Goal Recognition}
{Alshehri, Abdulrahman, Alamri, Miller \& Vered}

\firstpageno{1}

\begin{document}

\title{Towards Explainable Goal Recognition Using Weight of Evidence (WoE): A Human-Centered Approach}

\author{
\name Abeer Alshehri \email aalshehri@student.unimelb.edu.au \\
\addr School of Computing and Information Systems, The University of Melbourne\\ 
Melbourne, Australia\\
Department of Computer Science and Information Systems, King Khalid University\\
Abha, Saudi Arabia
\AND
\name Amal Abdulrahman \email amal.abdulrahman@mq.edu.au \\
\addr School of Computing, Macquarie University\\
Sydney, Australia
\AND
\name Hajar Alamri \email hajmohemmad@kku.edu.sa  \\
\addr Department of Computer Science and Information Systems, King Khalid University\\
Abha, Saudi Arabia
\AND
\name Tim Miller \email timothy.miller@uq.edu.au \\
\addr School of Electrical Engineering and Computer Science, The University of Queensland\\
Melbourne, Australia
\AND
\name Mor Vered \email mor.vered@monash.edu  \\
\addr School of Computing and Information Systems, Monash University\\
Melbourne, Australia
}


\maketitle

\begin{abstract}

Goal recognition (GR) involves inferring an agent's unobserved goal from a sequence of observations. This is a critical problem in AI with diverse applications. Traditionally, GR has been addressed using 'inference to the best explanation' or abduction, where hypotheses about the agent's goals are generated as the most plausible explanations for observed behavior. Alternatively, some approaches enhance interpretability by ensuring that an agent's behavior aligns with an observer's expectations or by making the reasoning behind decisions more transparent. In this work, we tackle a different challenge: explaining the GR process in a way that is comprehensible to humans. We introduce and evaluate an explainable model for goal recognition (GR) agents, grounded in the theoretical framework and cognitive processes underlying human behavior explanation. Drawing on insights from two human-agent studies, we propose a conceptual framework for human-centered explanations of GR. Using this framework, we develop the \textit{eXplainable Goal Recognition} (XGR) model, which generates explanations for both \textit{why} and \textit{why not} questions. We evaluate the model computationally across eight GR benchmarks and through three user studies. The first study assesses the efficiency of generating human-like explanations within the Sokoban game domain, the second examines perceived explainability in the same domain, and the third evaluates the model's effectiveness in aiding decision-making in illegal fishing detection. Results demonstrate that the XGR model significantly enhances user understanding, trust, and decision-making compared to baseline models, underscoring its potential to improve human-agent collaboration.

\end{abstract}

\section{Introduction}

Goal Recognition (GR) is the problem of predicting an agent’s intent by observing its behavior. The task of GR has numerous potential and practical applications, such as smart homes \cite{hegde2019survey} and workplace safety \cite{inam2018risk}, among others \cite{wayllace2020dragon,singh2020combining}. Research on GR uses different inference techniques to predict the ultimate goals of the agents being observed. It advances with increasingly complex domain models and better approaches. However, understanding and fostering human trust in these systems is challenging due to their lack of explainability. This becomes particularly crucial in safety-critical applications like social care, military planning, and medical support, where the system's decisions can have major consequences. Systems must be capable of explaining the decisions made and communicating their decisions in a way that is understandable to people \cite{masters2021s,meneguzzi2021survey,van2021activity}.

The vast majority of work has focused on improving the explicability of agent behavior \cite{yolanda2015fast,sohrabi2016plan,vered2016online,Hu2021-ds,Hanna2021-rm}. This typically involves making the behavior more understandable to observers, either by aligning it with their expectations or ensuring the interpretability of the inference process. In shifting the focus from interpretability to justification, the goal is to develop an explainable GR agent that provides context and rationale for each predicted goal, rather than merely making the inference process interpretable.

This paper contributes to the ongoing work in eXplainable AI (XAI) by developing an explainable GR model from a human-centered perspective. Studies in social psychology indicate that people use the same conceptual framework that they apply to humans to explain artificial agents' behavior, and they also expect artificial agents to adopt this framework \cite{de2017people}. Therefore, our approach is to provide comprehensible explanations that align with how people rationalize outcomes and interpret information. This effort aims to bridge the gap between mere prediction and understandable explanation.

There are two main contributions of this work. First, we propose a conceptual framework for the explanation process in GR tasks and identify key concepts people use when reasoning about goal prediction. Using a bottom-up approach, the framework is derived from an analysis of human explanations of recognition tasks. We study this in two different domains to increase the generalizability of our model: Sokoban and StarCraft games. We examine the frequency, sequence, and relationships between the basic components of these explanations. Using the thematic analysis process, we identified 11 concepts from 864 explanations of agents operating in various scenarios. Incorporating insights from the folk theory of mind and behavior \cite{malle2006mind}, we propose a human-centered model for GR explanations.

Our second contribution is building an \textit{eXplainable Goal Recognition} (XGR) model based on the proposed conceptual framework. The model generates explanations for GR agents using the information theory concept of Weight of Evidence (WoE) (Good 1985; Melis et al. 2021). We define the problem of explanation selection using the main concept from our conceptual model, which we call an \emph{observational marker}, i.e., the observation with the highest WoE. We computationally evaluate the XGR model on eight GR benchmark domains \cite{vered2018towards}. We conducted three user studies to evaluate our model's performance in different aspects: 1) Efficiency in Generating Human-Like Explanations: This study focused on assessing how well the model produces explanations that resemble those generated by humans, specifically within the Sokoban game domain; 2) Perceived Explainability: This study examined how users perceive the model’s ability to explain its actions and decisions, also within the Sokoban game domain. 3) Effectiveness in Supporting Decision-Making: This study evaluated the model's effectiveness in supporting users' decision-making process in the domain of illegal fishing detection. In the first study, our model aligns with human explanations in over 73\% of scenarios. In the second and third studies, our model outperforms the tested baselines.

Part of this paper was published at the International Conference on Automated Planning and Scheduling (ICAPS) \cite{alshehri2023explainable}, where we presented our XGR model and its evaluation through the first two user studies. In this work, we introduce the conceptual framework for GR explanations, provide further details on our XGR model, and extend its evaluation to the context of decision-making support by conducting a third user study.

The structure of the paper is as follows. Section 2 reviews the related work on explainability in GR and human behavior explanation; Section 3 presents the human-agent experiment to build the conceptual framework; Section 4 provides the necessary background required to follow up with the proposed XGR model; Section 5 presents the XGR model; Section 6 describes experiments that evaluate the model; We then conclude with a summary and opportunities for further research in Section 7.


\section{Related Work}

\subsection{Goal Recognition and Explainability} 

Goal recognition (GR) involves identifying an agent's unobserved goal based on a sequence of observations. Various approaches exist to address the GR problem. Common methods include library-based GR algorithms, which use specialized plan recognition libraries to represent all known methods for achieving known goals \cite{sukthankar2014plan}; model-based GR algorithms \cite{ramirez2010probabilistic,sohrabi2016plan,vered2016online}, where GR agents leverage domain knowledge through planners to generate the necessary plans for achieving a goal \cite{masters2021s}; and machine learning GR approaches that rely on large training datasets from which algorithms learn domain constraints \cite{min2014deep,pereira2019online,meneguzzi2021survey,fitzpatrick2021behaviour}. Well-established algorithms have shown high performance in labeling action sequences with corresponding goals \cite{ramirez2010probabilistic,vered2018towards,pereira2020landmark}, yet explaining why the algorithm arrived at a particular conclusion remains under-explored.

Prior work has focused on explaining GR in the form of answering the question: what goal is the agent trying to achieve? A long line of work has suggested explaining goal inference, which is a form of 'inference to the best explanation', also called abduction \cite{van2008goals,baker2008theory,baker2012bayesian,blokpoel2013computational,zhi2020online}. This process formulates hypotheses about the agent's goals which are identified as the most plausible explanations for the observed actions. That would assist in making sense of actions and attributing appropriate goals or intentions to them, enhancing our understanding of the agent's behavior. Another approach is improving the explicability of agent behavior 
\cite{yolanda2015fast,sohrabi2016plan,vered2016online,farrell2020narrative,Hu2021-ds,Hanna2021-rm,DBLP:journals/corr/abs-2302-09646}. This involves ensuring that the behavior of the agent is self-explanatory to an observer by either aligning its actions with the observer's expectations or making the reasoning behind its decisions interpretable. These approaches often assume optimal or simplified sub-optimal actions, neglecting the inherent challenges in agent planning. \citeauthor{keren2014goal} \citeyear{keren2014goal} introduced the Goal Recognition Design (GRD) approach, which facilitates the process of inferring an agent's goals. This approach aims to analyze and redesign the underlying domain environment to ensure early and accurate detection of the agent's objectives.

However, previous approaches assume that the agent's behavior or domain is controlled to make its actions explicable. In this work, we address a different problem: explaining the GR process in a way that is understandable to humans. In complex and dynamic environments where agent behavior may not neatly align with human expectations or optimal action models, understanding the reasoning behind goal predictions is crucial for building trust in GR systems. There is a need for an explanation model that justifies the GR output, ensuring that the reasons behind goal predictions are clear and understandable. Instead of merely making the inference process transparent by controlling the domain or agent's actions, a model should provide context and rationale for each predicted goal. This includes accounting for sub-optimal behavior that might arise due to planning difficulties or environmental constraints, thereby offering a more nuanced and realistic understanding of agent actions.

Additionally, fostering a solid understanding of an agent's behavior presents a significant challenge for decision-makers. Human-AI team performance is influenced by scenarios where the AI system provides predictions while humans maintain the final decision-making authority. GR systems play an essential role in this context by accurately predicting and interpreting user intentions, which inform and guide subsequent actions \cite{pushp2017cognitive,ognibene2019proactive,brewitt2021grit,jamakatel2023towards}. While these systems demonstrate proficiency in high-stakes event prediction, they often lack the ability to provide justifications that clarify the motivations behind predicted intentions. The need for a human-like explanation should be considered to elevate system prediction toward a cognitive understanding of why certain outcomes are predicted and how they relate to the broader context of high-stakes situations \cite{sahoh2023role}. It will enhance decision-making by ensuring that AI predictions are effectively understood and appropriately used.

\subsection{Human behavior explanation} 

We outline work on social attribution, which defines how people attribute and explain others' behavior. Social attribution focuses not on the actual causes of human behavior but on how individuals attribute or explain the behavior of others. \citeauthor{Heider1958THENA} \citeyear{Heider1958THENA} defines social attribution as person perception, emphasizing the importance of intentions and intentionality. An intention is a mental state where a person commits to a specific action or goal. People consistently agree on classifying events as either "intentional" or "unintentional" \cite{malle1997folk}. It is argued that while intentionality can be objective, it is also a social construct, as people ascribe intentions to one another, impacting social interactions.

In addition to intentions, research suggests that other factors, such as beliefs, desires, and traits, play a significant role in attributing social behavior. Researchers from various fields have converged on the insight that people's everyday explanations of behavior are rooted in a basic conceptual framework, commonly referred to as folk psychology or theory of mind \cite{Heider1958THENA,horgan2013folk,10.1093/acprof:oso/9780195307696.003.0010}. Folk psychology involves attributing human behavior using everyday terms like beliefs, desires, intentions, emotions, and personality traits. This area of cognitive and social psychology acknowledges that, although these concepts may not genuinely cause human behavior, they are the ones people use to understand and predict each other's actions \cite{malle2006mind}.

\begin{figure}[ht]
\includegraphics[width=9cm]{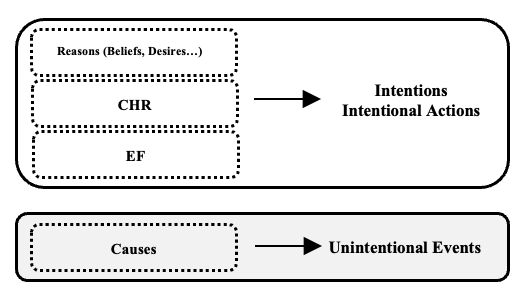}
\centering
\caption{Malle's conceptual framework for behavior explanation; adapted from \cite{de2017people}}
\label{figure0}
\end{figure}

Malle [110] presents a model grounded in the Theory of Mind to explain how people attribute behavior to others and themselves by assigning mental states such as desires, beliefs, values, and intentions. This model identifies different modes of behavior explanations and their cognitive processes by distinguishing between intentional and unintentional actions (Figure \ref{figure0}). Intentional behavior is typically explained by reasoning over key mental components—the reasons behind deliberate acts—based on the rationality principle, where agents are expected to act efficiently to achieve their desires given their beliefs and values. Sometimes, people explain intentional actions using two additional modes: causal history of reason explanations (CHR) and enabling factor explanations (EF). In CHR mode, people focus on factors that influence the reasons behind an action, such as unconscious motives, emotions, and cultural influences, which do not necessarily involve rationality or subjectivity. In EF mode, instead of explaining the intention, they explain how the intention led to the outcome, considering personal abilities or environmental conditions that facilitated the action. Unintentional behavior, on the other hand, is explained by referring to causes like habitual or physical phenomena, particularly those that prevent or prohibit intentional behavior.

Our proposed model is based on Malle's framework, focusing on the mode of reasoning for intentions and intentional actions. Since people tend to attribute human-like traits to artificial agents, they expect explanations from these agents using the same conceptual framework \cite{de2017people}. Therefore, we develop our model by incorporating insights from human-agent studies. To the best of our knowledge, this is the first model designed for explainable GR agents.



\section{Human-Agent Study: Insights from Human Explanation} \label{human studies}

In this section, we present our conceptual model for explaining GR, grounded on empirical data from two different human-agent studies.

\subsection{Study Objective}

 The goal of this study is to investigate how humans explain a GR agent's behavior and identify the key concepts present in such explanations, their frequency, and the relationships among these concepts. Based on our findings, we construct a conceptual framework for GR explanation. We have two case studies with different scenarios and assumptions. The first case study is set in a general domain where no specific expertise is required, such as the Sokoban domain, a classic puzzle game where a player pushes boxes to designated storage locations within a grid. In contrast, the second case study involves explanations provided by domain experts, as seen in the StarCraft domain, a complex real-time strategy game that requires strategic planning, resource management, and tactical combat.

\subsection{Goal Markov Decision Process (Goal MDP)}

For both case studies, we employ the Goal Markov Decision Process (Goal MDP) framework to capture an observer's view of the world. A Goal MDP \cite{ramirez2011goal} represents the possible actions that can be taken and the causal relationships of their effects on the world's states. Formally, it is defined as a tuple $\Pi = (S, S_{G}, A, P, C)$, where $S$ is a non-empty state space, $S_{G}$ is a non-empty set of goal states, $A$ is a set of actions, $P_a(s' \mid s)$ is the probability of transitioning from state $s$ to state $s'$ given action $a$, and $C(s, a, s')$ is the cost of that transition. The solution to a Goal MDP is a policy $\pi: S \rightarrow A$ that maps states to actions with an overall minimum expected cost.

We describe an observer's worldview as a Goal MDP. Our description is based on the following assumptions: (1) the observer perceives the world as a finite set of discrete states and actions; (2) the observer interprets transitions between states in a deterministic manner, while the probability reflects the observer's confidence or uncertainty about the transition; (3) the observer values actions based on costs, aiming to minimize them over time; (4) the observer's reasoning is based on their internal world state representation, which may not necessarily match the actual state observed by the player/agent; (5) the player's preferences are not observable, meaning the observer must infer information and rely on observable actions to make judgments; and (6) the observer is assumed to have full observability, meaning they have access to the complete state of the environment.

\subsection{Case Study 1: Sokoban Game} \label{sokoban}
\begin{figure}[ht]
\includegraphics[width=12cm]{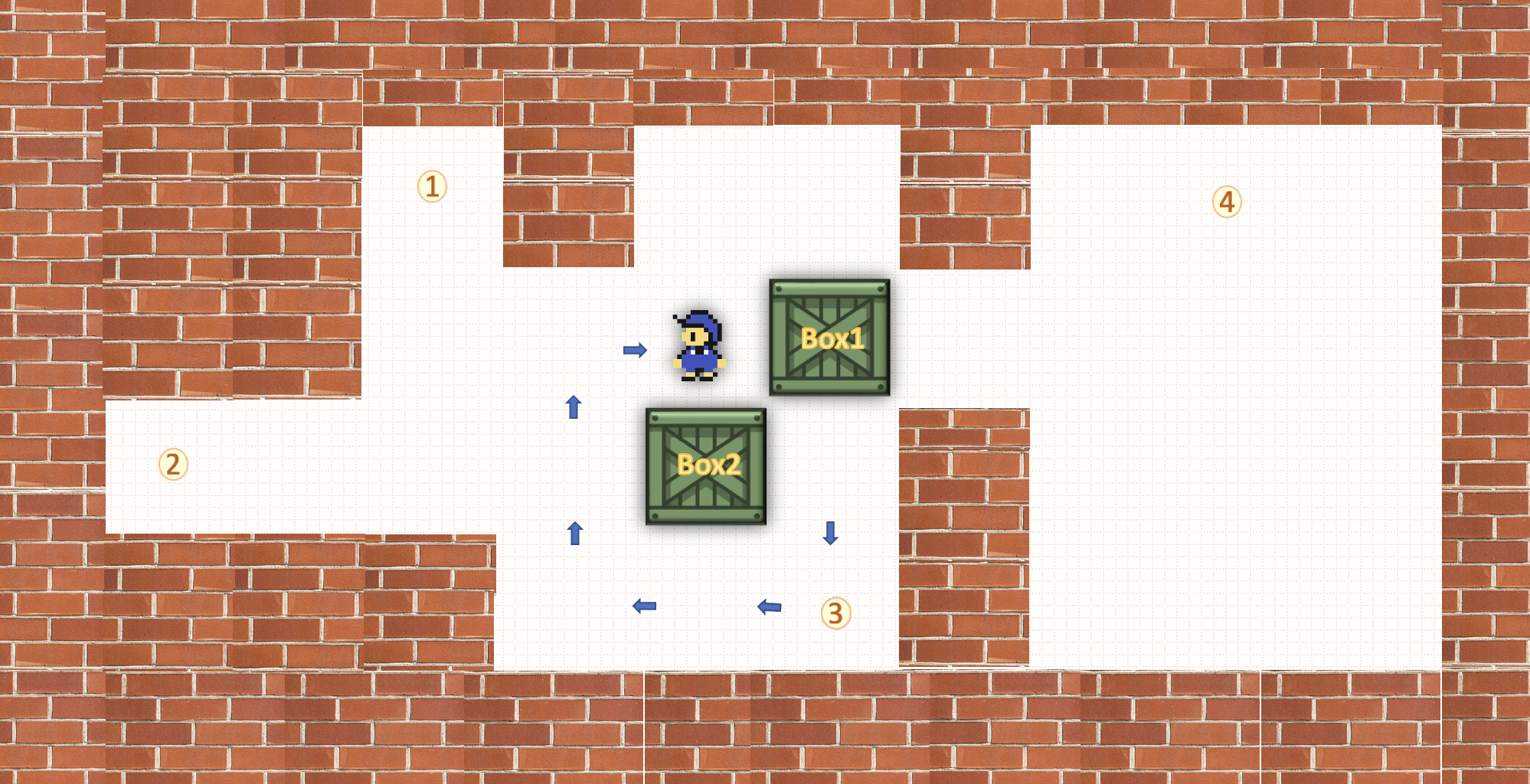}
\centering
\caption{Screenshot of the Sokoban game. There are four storage locations (marked as yellow circles). The possible goals involve delivering two boxes to any combination of these locations. Blue arrows represent observations of the player's move so far.}
\label{figure1}
\end{figure}

Sokoban is a classic puzzle game  (Figure \ref{figure1}) set in a warehouse environment, where the player or agent navigates through a grid-like layout to move boxes onto designated storage locations.  The objective is that each box must be pushed, one at a time, to its assigned spot. The challenge lies in navigating movement constraints and spatial limitations; boxes can only be pushed into empty spaces and cannot be pulled or pushed against walls or other boxes. We modified the Sokoban game rules to allow the player to push multiple boxes simultaneously. This modification transforms the game from a straightforward navigational task into a strategic challenge with multiple objectives, where the player aims to minimize the number of steps taken. We used a STRIPS-like discrete planner to generate plan hypotheses derived from the domain theory and observations as our ground truth. 

\subsubsection{Study Design}

We defined the observer's worldview within the framework of Goal MDPs as follows:

\begin{itemize}
\item 
State Space $S$: Represents the snapshot of the world, including the player's position, walls, boxes, and storage locations at any given point in time.
\item 
Action Space $A$: Encompasses the various actions the player can take, either moving or pushing the box(es) in one of the following directions: up, down, right, or left.
\item 
Cost Function $C(s, a, s')$: Assigns a cost for each action to encourage making the fewest amount of moves.
\item
Goal State $S_{G}$: Includes all possible goals that the player can achieve within the current state of the world. Each goal represents a desired configuration of boxes in relation to storage locations.

\end{itemize}

Table ~\ref{table1} presents different goal representations based on the original game version. In versions 1 and 2, the player can push one box at a time, whereas in version 3, they can push two boxes at a time to goal locations. The goal recognition problems include a number of competing goal hypotheses: multiple possible plans to achieve a goal, two sequential goals with interleaved plans to achieve them, or failed/unsolvable plans (See Appendix A for screenshots of different scenarios).

\begin{table}[ht]
\scriptsize
\centering
\begin{tabular}{lp{65mm}}
\toprule
\textbf{Version}    & \textbf{Player Task}  \\ 
\midrule
Game 1 & Deliver one box to one of three possible goal locations; push one box at a time. \\[1mm]
Game 2 & Deliver two boxes to two of four possible goal locations; push one box at a time. \\[1mm]
Game 3 & Deliver two boxes to two of six possible goal locations; push multiple boxes at a time.\\ 
\bottomrule
\end{tabular}
\caption{Player game versions.} 
\label{table1}
\end{table}

We additionally varied scenarios to present different \textit{rationality levels}, including rational (optimal or suboptimal) behaviors and irrational behaviors. For optimal behavior, we assumed that observers would form a simple notion of optimal behavior, where the player takes the shortest path toward a goal. In the suboptimal behavior scenario, the player either chooses a longer path toward a goal (suboptimal plan) or deviates from a rational action in an observed sequence of a particular goal plan (e.g., the agent's goal may have changed). We also included irrational behaviors where the player fails to complete the task (e.g., getting stuck in a dead-end state) to observe how these actions would be interpreted by participants. 

The participant's task was divided into the following phases: 
\begin{enumerate}
  \item Watch an instructional video to introduce the task and game rules.

  \item For each of the 18 different scenarios (three games, six scenarios per game): 
  
   \begin{itemize}
       \item  Watch a video clip in which a player tries to achieve the task  (Figure~\ref{figure1}).

       \item After watching the observed actions sequence (plan's completion percentage $\textit{min} = 0.25\%, \quad \textit{median} = 0.53\%, \quad \textit{max} = 0.83\%$), 
       predict which goal location the player is trying to get to, and, accordingly, assign a likelihood (with one as the least likely and five as the most likely) of each goal. This prediction task is not central to the objectives of this study but was used to engage participants in reasoning about behavior.
  
       \item Provide reasons for your prediction. Participants were required to answer specific questions based on the condition they were in.
   \end{itemize}  

\end{enumerate}

Each participant was  randomly assigned to one of  three conditions: 
\begin{itemize}
    \item `Why' condition: participants were asked to: "Explain \textit{why} you have rated that/those goal(s) as the most likely?"
    \item `Why-not' condition: participants were asked to: "Explain \textit{why} you have not rated that/those goal(s) as the most likely?"
    \item `Dual' condition: participants were asked to explain \textit{both} \textit{why} and \textit{why you have not} in that order.

\end{itemize}

We collected data for the first and second conditions to analyze the differences between \emph{why} and \emph{why not}, and for the third condition to analyze how people answer \emph{why not} if they have already answered \emph{why}, and how the answer of \emph{why} differs if they know there is a \emph{why not}.


\subsubsection{Data}

We recruited 36 participants (22 male, 14 female), allocated evenly and randomly to each condition, aged between 20 and 65, with a mean age of 38. We limited the study to participants from the United States who are fluent in English. Recruitment was conducted via Amazon Mechanical Turk. Participants were compensated \$6.50 for completing the task and a bonus of \$3.50 for providing more thoughtful answers.

With three different game versions, six scenarios per game, and 12 participants per condition, a total of 864 textual data points were collected (Table \ref{table2}). We used several methods to filter out deceptive participants. We excluded explanations with fewer than three words or containing gibberish. We also used the time taken to complete the survey as a threshold. This left us with a total of 828 explanations.

We used participants' open-ended explanations to better identify the concepts they used to explain the player's predicted goal. The word count of given answers within the dataset is between 1 and 98 words (\(\bar{x}_1=22.96\), \(\sigma_1=15.52\)) for the first condition, 3 and 81 words (\(\bar{x}_2=26.57\), \(\sigma_2=15.63\)) for the second condition, and 1 and 64 words (\(\bar{x}_3=20.38\), \(\sigma_3=11.65\)) for the third condition.


\begin{table}[ht]
\scriptsize
\centering
\begin{tabular}{@{}lccc@{}c@{~}c@{}}
\toprule
 & \multicolumn{3}{c}{\#Questions per game} & \\ \cmidrule(lr){2-4}
Conditions & G1 &  G2 & G3  & Participants&  Explanations \\ \midrule
\textbf{Why}        & 6                                     & 6             & 6   & 12                                 & 216                           \\
\textbf{Why-not}        & 6                                    & 6                 & 6    & 12                                & 216                           \\ 
\textbf{Dual}        & 12                                   & 12                                   & 12     & 12                              & 432                           \\ \bottomrule
\end{tabular}
\caption{Data sources}
\label{table2}
\end{table}

\subsection{Case Study 2: StarCraft Game} 
\begin{figure}[ht]
\includegraphics[width=12cm]{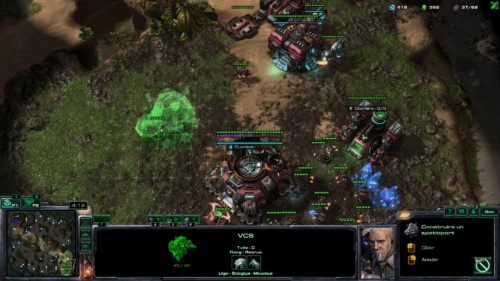}
\centering
\caption{Screenshot of the StarCraft game}
\label{figure2}
\end{figure}

StarCraft is a real-time strategy (RTS) game (Figure \ref{figure2}) where players manage an economy, produce units and buildings, and compete for control of the map with the ultimate aim of defeating all opponents. As an RTS game, StarCraft has several defining characteristics \cite{ontanon2013survey}: 
\begin{itemize}
    \item Players engage in a Simultaneous Move Game, where they execute actions concurrently—such as moving units, building structures, and managing resources—demanding effective multitasking skills;
    \item The Partially Observable Domain limits players' visibility of the game map and opponents' setups, necessitating strategic reconnaissance for informed decisions;
    \item Real-time gameplay adds urgency, requiring rapid thinking and reflexes to outmaneuver opponents within time constraints; 
    \item  Non-Deterministic Actions introduce uncertainty, challenging players to adapt strategies dynamically;
    \item The game's High Complexity stems from its vast state space and diverse strategic options involving units, buildings, and technologies, compelling players to consider multiple factors when planning strategies. 
\end{itemize}  

These elements combine to create a deeply strategic and challenging gaming experience, where success depends on a player's ability to react at strategic, economic, and tactical levels.

During gameplay, shoutcasters (commentators) in esports deliver real-time explanations and commentary to audiences, elucidating the intricate strategies and tactics of the game to enhance accessibility and engagement. Their expert insights can potentially benefit XAI tools by analyzing their commentary in the RTS domain \cite{penney2021shoutcasters}.

\subsubsection{Study Design}
We defined the worldview of shoutcasters, i.e. observers,  within the framework of Goal MDPs as follows:
\begin{itemize} 
\item State Space $S$: Encapsulates the game environment's configurations and conditions at any given time. This includes the positions of units, resources, and other relevant game elements. 

\item Action Space $A$: Includes low-level actions of specific game units and high-level actions related to game strategies and tactics (sequences of actions). 
\item Cost Function $C(s, a, s')$: Quantifies the value of states and actions in terms of progress toward winning. 

\item Goal State $S_{G}$: Comprises both subgoals (intermediate objectives that players or teams aim to achieve) and the main goal (the ultimate objective, typically winning the game). 
\end{itemize}
We analyzed a dataset comprising professional StarCraft tournament videos \cite{penney2021shoutcasters}, where expert shoutcasters provided commentary. We identified key instances where shoutcasters made predictions about players' goals and strategies and coded these instances to capture the underlying concepts used in their explanations. We clustered the dataset to ensure representative sampling.

\subsubsection{Data}

We obtained the dataset from \citeauthor{penney2021shoutcasters} \citeyear{penney2021shoutcasters}, which was collected from professional StarCraft tournaments available as videos on demand from 2016 and 2017. They selected 10 matches and then randomly chose one game from each match. Each of the 10 videos features two shoutcasters (expert commentators) providing commentary.

To obtain representative samples, we identified six clusters within the dataset (1387 instances divided by 6 clusters equals approximately 231 instances per cluster). We then randomly selected one sample of 50 instances from each cluster. The resulting sample size was 300 instances (6 clusters * 50 instances each). We only considered instances involving predictions, specifically when shoutcasters explain their recognition process of what the agents/players aim to achieve (their goals) and so ended up having a total of 132 instances out of the six samples.  As the data source is public, we provided supplementary material of coded data to support future research.

\subsection{Method}

We used a hybrid approach of deductive and inductive reasoning, employing thematic analysis as outlined by \citeauthor{braun2006using} \citeyear{braun2006using} to analyze our data. The analysis process was divided into six phases: familiarization with the collected data, development of codes, sorting different codes into potential themes, reviewing themes, defining and naming themes, and writing up the report.

Initially, the collected data was re-read multiple times to ensure immersion before proceeding to the coding phase. The coding process was inductive, aiming to identify basic concepts regarding how people explain goals, and deductive by relating them to the existing literature on explaining human behavior \cite{malle2006mind}. Malle's explanation model \cite{malle2006mind} shows that people reason over others' beliefs, values, and desires to explain intentional actions. Following that model, we apply these concepts to our coded data. Subsequently, the codes were grouped into defined themes based on their similarities.

In the context of this study, the proposed themes are linked to our research topic of explaining human behavior in goal recognition scenarios. After establishing a set of candidate themes, the refinement process focused on ensuring internal homogeneity—i.e., a cohesive pattern within each candidate theme to accurately reflect the overall data set. Relationships, links, and distinctions between themes were identified during this phase.

The next steps involved naming and describing the themes and illustrating the thematic elements with examples. To ensure consistency, four authors independently coded 10\% of the data, achieving approximately 75\% inter-rater reliability, as measured by percentage agreement. After achieving this level of agreement, the first and fourth authors continued to code the remainder of the data.


\subsection{Results}

\begin{figure}[ht]
\includegraphics[width=13cm]{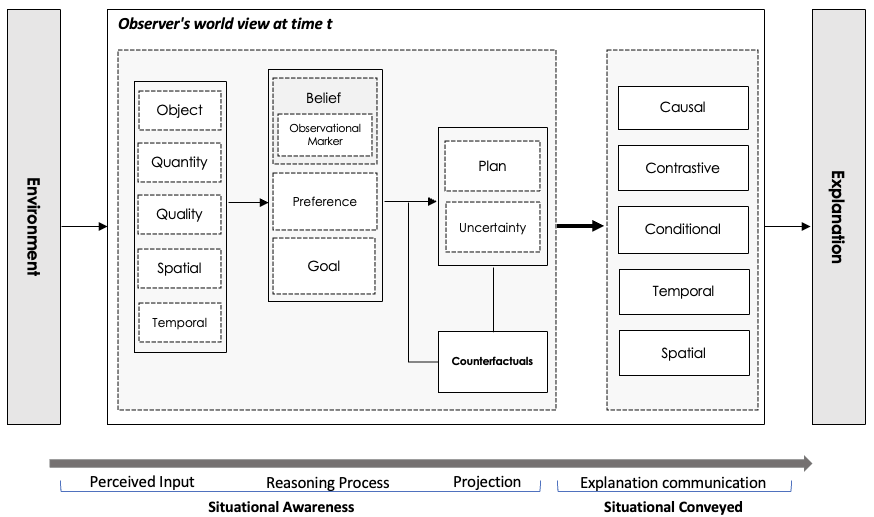}
\centering
\caption{The conceptual model of explainable goal recognition} 
\label{figure3}
\end{figure}

\subsubsection{The Conceptual Model of Goal Recognition Explanation}

We developed the conceptual model by integrating insights from both studies. Initially, we built a model based on the findings from the Sokoban study. This model was then extended to encompass the complexities of the Starcraft study, ensuring that it accurately reflects both studies' dynamics and unique aspects. We present our conceptual model of an explainable Goal Recognition (GR) agent in Figure \ref{figure3}. The figure highlights the common elements (concepts) that encode an observer's view of the world and the different representations of given explanations. The model process is guided by two levels; situational awareness and situational conveyance. The situational awareness model, presented by Endsley's in  \cite{endsley1995toward}, is a widely used situation awareness model consisting of three consecutive stages:

\begin{enumerate} 
\item \textbf{Perceived input:} In this stage, the observer perceives the basic elements and their properties in the environment—object, quantity, quality, spatial and temporal information.

\item \textbf{Reasoning process:} Based on the perceived inputs, the observer makes causal inferences between their beliefs (including the actor's mental model) and goals, generates counterfactuals, and associates them with their preferences.

\item \textbf{Projection:} This stage presents the observer's predictions of future actions guided by their expected goals and the uncertainty level based on the understanding of the previous stages. 
\end{enumerate}

In this level, two types of reasoning occur due to the two mental model representations \cite{felli2015computing}. The first type is stereotypical reasoning, in which the observer reasons about others' mental states (what the observer would have done). An example from the data corpus is: ``It might just do old classic seven gate [a game strategy]." The second type is empathetic reasoning, where the observer casts themselves into the actor's mental model and reasons as they would (what the actor would have done). An example is: ``This is exactly what I was talking about, you do something to try to force them." In practice, the observer often contrasts both views in a single explanation—what they think should happen vs. what the actor is likely to do. For example: ``It's actually going to look for a run by here with this scan it looks like but unfortunately unable to find it with the ravager here poking away." This was observed when explanations were provided for either a failed plan or a sub-optimal one toward achieving a goal. For simplicity, we assume a local perspective in this work, where the actor's mental state is equal to the observer's through the data coding and model implementation process.

The second level is a situational conveyance, where different explanations are formed and communicated by the observer. At this level of the model, the explaining process requires additional strategic knowledge. This includes reasoning about contrastive, conditional, temporal, and spatial cases of problem-solving tasks, allowing for more than just a causal representation of a given explanation.


\begin{table}[ht]
\centering
\small 
\caption{Codes (of the concepts) and their descriptions of explanations across two human studies, with examples given from different participants. }
\label{table3}
\begin{tabularx}{\textwidth} { 
  >{\raggedright\arraybackslash}X 
  >{\raggedright\arraybackslash}X 
  >{\raggedright\arraybackslash}X }
\toprule
Code                & Description                           & Example                         \\ \midrule
Causal              & A cause and effect relationship         &      ``\textbf{because} it was positioned closer''        \\
Conditional         & A conditional relationship               &       ``\textbf{if} jjakji takes a 3rd''      \\ 
Contrastive         & A contrastive relationship               &     ``blocked for any goal \textbf{but} 1''        \\
Temporal            & A series of events over time (order, repetition, opportunity chain, timing)  &        ``a very strong timing that he can hit where he might \textbf{be able to} kill'' \\
Spatial             & Refer to places or distances                      &        ``he has moved \textbf{above} it''     \\
Preference/Judgement           & Assessment of actions or outcomes      &       ``he gets the \textbf{perfect} split''      \\
Goal                & Refer to a goal state(s)                      &        "it likely wanted to go to \textbf{goal 1}"     \\
Plan                & Refer to a future action or sequence of actions    &     ``he is \textbf{going to snipe} the warp prism''     \\
Counterfactual goal & Refer to counterfactual goal(s) that could have occurred under different conditions        &     ``boxes have been moved away from \textbf{goal one}''       \\
Counterfactual Plan & Refer to a counterfactual action or sequence of actions            &   ``instead of \textbf{building a robo in a prism}''  \\
Belief              & Refer to general domain knowledge               &        ``\textbf{No vision} on the left-hand side of the map''      \\
Observational Marker &  Refer to the observed precondition(s) that supports the hypothesized goal(s)   &    ``Given \textbf{the player's last move}, box 1 belongs on goal 3''   \\
Counterfactual Observational Marker  & Refer to the observed precondition(s) that is against the hypothesized counterfactual goal(s)   &   ``The player \textbf{would have taken different steps} if position 1 was the goal''   \\
Object              & An object in the domain                                  &   ``to build a \textbf{gate}'' \\
Quantitative        & Refer to some quantity or measured value of an object   &   ``a turret at \textbf{90\% }complete''    \\
Qualitative         &  Refer to some quality or characteristic of an object    &     ``here with \textbf{inferior} roaches'' \\
Uncertainty         & Refer to a state of being uncertain                      &    ``I \textbf{suspect} it is to push the block to goal 2''  \\ \bottomrule
\end{tabularx}
\end{table}

\subsubsection{Concepts} \label{concepts}

Table \ref{table3} shows the different codes and concepts that emerged from across the two studies. The given explanations include factual and experiential knowledge (`belief'), subjective likes and objective assessments (`preference'), the desired state to be achieved (`goal'), and possible future actions (`plan'). When people explain others' actions, they infer their goals to provide better explanations \cite{mcclure1997you}. When explaining a recognized goal, people infer the most relevant evidence from their belief state. Thus, we break down the belief concept into an `observational marker', an observed precondition that most influences goal prediction. This concept applies not only to optimal behavior, measured by traditional efficiency metrics such as time and shortest route, but also to suboptimal behavior.

Since recognition problems activate counterfactual thinking \cite{epstude2008functional},  explanations of GR reasoning also include `counterfactuals' — observational markers, plans, and goals. Explaining counterfactual plans implies having counterfactual goals where the actor has no plan to achieve them. We introduced an uncertainty code as we found that observers use words expressing uncertainty to indicate their confidence level. Additionally, as the problem is to explain goals, `goal' and `counterfactual goal' codes were also included. Finally, our data show the presence of different reasoning processes in the explanations. Observers associate a simple causal, conditional, contrastive, temporal, or spatial relationship when generating explanations. A combination of different concepts was used to form the given explanations.

Given that we conducted two distinct studies, we developed specific codes tailored to each. In the context of the Sokoban game, the code `counterfactual observational marker' is explicitly used in the explanations. This code refers to observed preconditions that are against the counterfactual goals. However, observers in the StarCraft game did not incorporate this concept in their explanations, likely due to time constraints and the assumed expertise level of the StarCraft audience.

In the StarCraft game, observers added additional object properties, such as quality and quantity, to describe the characteristics and measured values of objects. This detailed level of description is attributed to the greater complexity of the StarCraft game compared to Sokoban. Furthermore, observers included general domain knowledge (referred to as "belief") in their explanations to clarify facts that were inaccessible to the audience due to the partial observability of the game domain.

\begin{figure}[ht]
\includegraphics[width=13cm]{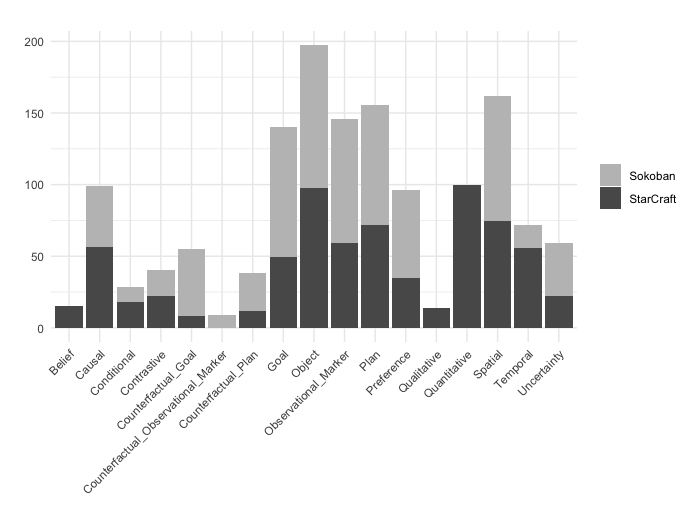}
\centering
\caption{Codes and their frequencies (\%) across the two human studies} 
\label{figure4}
\end{figure}

\subsubsection{Frequencies} \label{freq}

Figure \ref{figure4} illustrates the frequency distribution of 17 codes across two case studies. Given the navigational nature of both domains, observers predominantly referenced objects and their spatial properties to reflect the players' strategies.

Among the various codes, the observational marker emerges as the most significant finding. This code, which ranks fourth in frequency, provides crucial insights into how observers infer players' intentions and strategies. For instance, in the Sokoban game dataset, an observer noted, "The player positioned itself on top of the box, leading me to believe it is going to push down on the box to reach goal 2." Here, the action "positioned itself on top of the box" is used to explain the entire observed sequence, forming a critical precondition for achieving the predicted goal. This demonstrates how a single observed action can be pivotal in understanding the player's overall strategy. The counterfactual observational marker was coded only in the Sokoban game, as participants tended to use it when responding to 'why not' questions. In the dynamic and time-constrained environment of live StarCraft commentary, shoutcasters were observed to focus heavily on observational markers in their explanations, even when addressing 'why not' questions.

In addition, the third most frequently occurring code is the `plan' code, where the actor's goal is explained in terms of how future actions (plans) contribute to achieving that goal. This explanation, provided in terms of future actions, offers insights into potential actions and their execution \cite{norling2009modelling}. We believe that people tend to explain in terms of future actions as a way to resolve uncertainty. Causality is also frequently involved, as observers often associate causal relationships to generate explanations.

\begin{figure}[ht]    
\centering    
\subfloat[\centering Answering Why Question]{{\includegraphics[width=7cm]{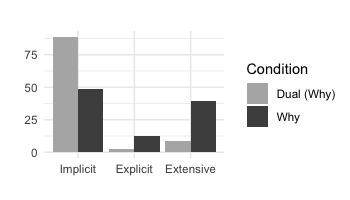} }}%
\qquad   
\subfloat[\centering Answering Why Not Question]{{\includegraphics[width=7cm]{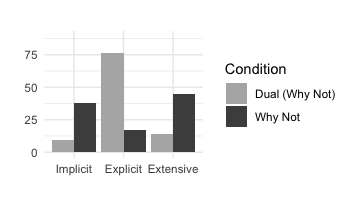} }}  
\caption{Explanation Modes and their frequencies (\%) over three conditions: Why, Why Not, and Dual.}%
\label{figure5}%
\end{figure}

\subsubsection{Questions} \label{questions}
In human studies, two forms of causal reasoning are used to answer certain questions \cite{hoffman2017explaining}: retrospective reasoning involves explaining past events through counterfactual reasoning, which considers what could have happened if the observed facts were different, prospective reasoning involves explaining future events through transfactual reasoning, which considers what could happen in the future if certain conditions are met. 

In the Sokoban game, we asked two questions: `Why?' and `Why Not', since it has been proved that they are the most demanded explanatory questions \cite{lim2009and}. "Why" questions typically demand contrastive explanations, which are addressed through counterfactual reasoning \cite{miller2019explanation}. In such explanations, people answer `Why A’ in the form of `Why A instead of B?’, where B is some counterfactual goal(s) that did not happen. From the data collected, we classified the provided contrastive explanations into three categories: \emph{implicit}, where observers implicitly contrast and identify relevant causes for A (the predicted goal(s)); \emph{explicit}, where observers explicitly contrast and identify relevant causes for B (the counterfactual goal(s)); and \emph{extensive}, where observers provide explanations for both by identifying relevant causes for A and also for B. It is important to note that the observer answered a ``why" question in the `why' condition, a ``why not" question in the `why not' condition, and both questions sequentially in the 'dual' condition. 

Figure \ref{figure5} illustrates the differences between conditions. In the dual condition, observers tended to adopt an implicit mode when they answered the why question--after having answered why-not—more frequently compared to the why' condition. A similar trend was observed for the why-not question in the dual condition, where observers tended to adopt an explicit mode when they answered the why-not question--after having answered why —more frequently compared to the Why-not condition. The contrastive nature of these explanations becomes particularly evident in the 'dual' condition, where observers can differentiate between the two types of questions. 

Participants primarily engaged in transfactual reasoning when they were highly uncertain about the agent's goal. This uncertainty is a result of the agent's observed behavior being suboptimal (irrational) to all goal hypotheses. For example, from the dataset: "If the player keeps pushing the two boxes together, it would be impossible for box 2 to be put back onto a goal."

The key codes highlight the important concepts used to explain goal recognition. Additionally, we coded implicit questions that shoutcasters tried to answer through their predictions. 

\begin{figure}[ht]
\includegraphics[width=0.6\textwidth]{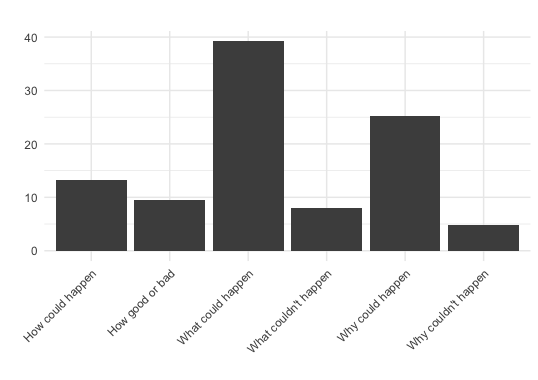}
\centering
\caption{Questions and their frequencies (\%) in StarCraft game} 
\label{figure6}
\end{figure}

In the StarCraft game, there are no pre-specified questions for the observers (shoutcasters) to answer. Instead, they gather information and craft explanations to address the audience's implicit questions. We coded implicit questions that shoutcasters tried to answer through their predictions. Table \ref{figure6} shows these questions and their frequencies. The shoutcasters are primarily focused on reasoning prospectively, addressing the "What could happen?" question allows them to anticipate future events. Despite the time constraints, they managed to provide reasoning for their predictions by answering other questions of interest.

The most frequently answered question is the ``Why?" question, involving retrospective reasoning over past events that mostly influence goal prediction. They further support their predictions by anticipating factors that control the context, allowing them to prospectively answer the ``How?" question through future projections.

\subsubsection{Discussion}

Causal reasoning is essential for constructing mental representations of events \cite{pearl2000models,malle2006mind,luo2010toward}. These representations form causal chains that illustrate how a sequence of causes leads to an outcome. In the realm of explainable AI, an agent aiming to explain observed events may need to use abductive reasoning to identify a plausible set of causes \cite{MILLER20191}. While numerous causes can contribute to an event, individuals typically select a subset they deem most relevant for their explanation \cite{MILLER20191}.

Our human studies focus on understanding how people explain others' predicted goals based on observed behavior. By coding these explanations, we identify a key concept, referred to as an 'observational marker' that participants use to build their explanations. Our findings align with social and cognitive research indicating that people prefer explanatory causes that seem sufficient in the given context for the event to occur \cite{spellman1997crediting,lipton1990contrastive,woodward2006sensitive,lombrozo2010causal}.

People make causal inferences about others' beliefs and goals based on their observed behavior and prior domain knowledge \cite{baker2009action}. A key aspect to explain those inferences is the ability to decide to what degree the observed evidence from a causal chain supports a goal hypothesis. To this end, we propose an explanation model for GR agents based on concepts such as causality and observational markers.

\section{Preliminaries}
In this section, we provide the essential background needed to follow the rest of the paper.

\subsection{Planning}
Planning aims to find a sequence of actions given an environment model, a current situation, and the goal to be achieved \cite{geffner2022concise}. The concept of planning is key to understanding GR algorithms that use planners in the recognition process.  We build upon the following planning problem definition as defined in \cite{pereira2017landmark}:

\begin{definition}
A planning task is represented by a triple $\langle \Xi, I, g \rangle$, in which $ \Xi = \langle F, A \rangle$ is a planning domain definition that consists of a finite set $F$ of facts that define the state of the world, and a finite set $A$ of actions; I is the initial state, and $g$ is the goal state. A solution to a planning task is a plan $\pi$ that reaches a goal state $g$ from the initial state I by following transitions defined in $\Xi$. Since actions have an associated cost, we assume that this cost is 1 for all actions.
\end{definition}

\subsection{Goal Recognition (GR)}
Goal recognition (GR) involves identifying an agent's goal by observing its interactions within an environment \cite{sukthankar2014plan}. We consider GR definition as defined by \cite{shvo2020active}.

\begin{definition}
\label{defn:GR}
A goal recognition problem is a tuple $\langle \Xi, I, G, O \rangle$, in which $ \Xi = \langle F, A \rangle$ is a planning domain definition where $F$ and $A$ are sets of facts and actions, respectively; $I$ is the initial state; $G = \{ g_{1}, g_{2}, ..., g_{m}\}$ is the goals set, and $O = \langle o_{1}, o_{2}, ..., o_{n}\rangle$ is a sequence of observations such that each $o_{i}$ is a pair $(\alpha_{i}, \phi_{i})$  composed of an observed action $\alpha_{i} \in A$ and a fact set that represent the state $\phi_{i} \subseteq F$. A solution to a GR problem is a probability distribution over $G$ giving the corresponding likelihood of each goal, i.e. the posterior probability $P(g_{j} \mid O )$ for each $g_{j} \in G$. The most likely goal is the one whose generated plan “best satisfies” the observations. 
\end{definition}

\subsubsection{The Mirroring GR Algorithm}
We focus on providing explanations for the output of the \textit{Mirroring} GR algorithm \cite{inproceedings,kaminka2018plan}. However, our approach is agnostic of the underlying GR algorithm and will work for any GR algorithm that fits Definition~\ref{defn:GR}. The \textit{Mirroring} algorithmis inspired by people's ability to perform online GR, originating from the brain's mirror neuron system, which is responsible for matching the observation and execution of actions \cite{rizzolatti2005mirror}. The approach falls under the \textit{plan recognition as planning} GR methods \cite{ramirez2010probabilistic,masters2021s} and uses a planner within the recognition process to compute alternative plans.

Specifically, the Mirroring algorithm uses a planner to compute optimal plans from an initial state $I$ to each goal $g_{j} \in G$ and to compute \textit{suffix} plans from the last observation $o_{i} \in O$ to each goal $g_{j} \in G$. Observations are processed and evaluated incrementally. These \textit{suffix} plans are then concatenated with a \textit{prefix} plan (the observation sequence $O$ at time step $t$) to generate new plan hypotheses. The algorithm subsequently provides a likelihood distribution, i.e., posterior probabilities $P(g_{j} \mid O)$ for each $g_{j} \in G$ by evaluating which of the generated plans, incorporating the observations $O$, best matches the optimal plan. The following example illustrates The Mirroring approach to solving a GR task.

\begin{figure}[t]
\centering
\includegraphics[width=0.5\textwidth]{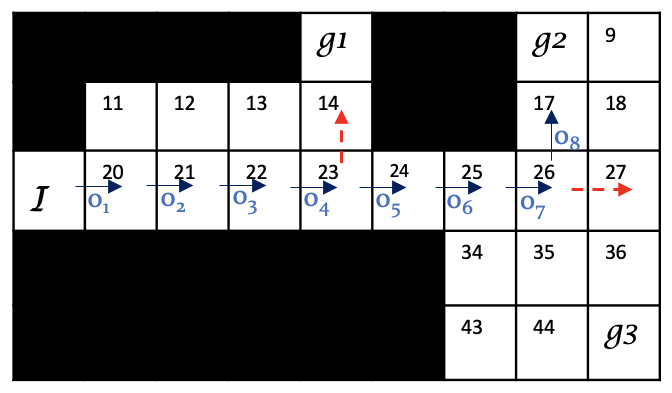} 
\caption{Navigational domain example.}
\label{figure7}
\end{figure} 

\begin{example} \label{running_example}
Figure \ref{figure7} presents a navigational domain where an agent can navigate through the unblocked grid to reach one of 3 possible goal locations. The GR task is composed of an initial state where an agent at the start is located (marked $
I$), a set of goal hypotheses, $G = \{g_1,g_2,g_3\}$, and a sequence of observations $O = \langle o_1,..,o_8 \rangle$ (represented as blue arrows). The domain definition $\Xi = \langle F, A \rangle$ includes a fact set $F$ comprising the cells (45 states in total) and an action set $A$ defined by four types of moves: up, down, left, and right, all with equal cost. The domain model is deterministic and discrete, meaning each action has only one possible outcome --- although our model does not assume deterministic actions. A goal state specification $G$ is defined as the agent being in one of the three possible goal cells. Thus, the Mirroring GR would infer that, most likely the agent's goal is to reach $g_2$ since the observed sequence confirms the optimal plan to achieve this goal.
\end{example}

\subsection{Weight of Evidence (WoE)}
The principle of rational action \cite{hempel1961rational} states that people explain goal hypotheses by assessing the extent to which each observed action contributes to a specific goal hypothesis over others. Building on this idea, \citeauthor{bertossi2020asp} \citeyear{bertossi2020asp} defines a causal explanation as the set of features most responsible for an outcome. By incorporating this approach, we model our explanation framework using the Weight of Evidence (WoE) concept.

Weight of Evidence (WoE) is a statistical concept used to describe the effects of variables in prediction models \cite{wod1985weight}. It is defined in terms of log-odds (see supplementary material) to measure the strength of evidence $e$ supporting a hypothesis $h$ against an alternative hypothesis $h'$, conditioned on additional information $c$. 
Assuming uniform prior probabilities\footnote{The derivation of the formula for non-uniform priors is provided in Appendix B.}, WoE is expressed as:
\begin{equation} woe(h/h' : e \mid c) = \log \frac{P(h \mid e, c)}{P(h' \mid e, c)}
\label{eq_woe} \end{equation}
Melis et al.\ \citeyear{melis2021human} propose a framework based on WoE for explaining machine learning classification problems, arguing that it aligns with how people naturally explain phenomena to one another \cite{miller2019explanation}. They found that WoE effectively captures contrastive statements, such as evidence for or against a particular outcome. This helps answer questions like why a goal $g$ is predicted, why not goal $g'$, and what should have happened instead if the goal is $g'$. We adopt this concept and apply it to GR problems.

\section{eXplainable Goal Recognition (XGR) Model}

Building on insights from the human studies discussed in Section \ref{human studies}, we propose a simple and elegant explainability model for goal recognition algorithms called \textit{eXplainable Goal Recognition} (XGR). This model extends the WoE framework to generate explanations for goal hypotheses. In the following section, we use the navigational GR example (Example \ref{running_example}) as a running example to support the definitions.

\subsection{Overview}
Extending \citeauthor{melis2021human}'s \shortcite{melis2021human} WoE framework, our model addresses `why' and `why not' questions, which are the most demanded explanatory questions \cite{lim2009and}. \citeauthor{lim2009and} \shortcite{lim2009and} compared a range of intelligibility-type questions and showed that explanations describing \textit{why} a system behaved in a certain way resulted in better understanding and increased trust in the system, which is also supported by our findings in Section \ref{questions} (Figure \ref{figure6}).

The model accepts four components as an input, which any GR model can provide:
\begin{enumerate}
\item The observed action sequence $O$.
\item The set of predicted goals, $G_p \subseteq G$.
\item The set of counterfactual (not predicted) goals, $G_c \subset G$, where $G_p \cap G_c = \emptyset$ and $G_p \cup G_c = G$.
\item The posterior probabilities $P(g \mid o_{i})$ and $P(g' \mid o_{i})$ for each $o_{i} \in O$, $g \in G_p$, and $g' \in G_c$.
\end{enumerate}

We define an explanation as a pair containing: (1) an explanandum, the event to be explained (i.e., goal hypotheses); and (2) an explanan, a list of causes provided as the explanation \cite{miller2019explanation}. The model answers two questions: `Why goal $g$?', where $g$ is a predicted goal hypothesis, and `Why not goal $g'$?', where $g'$ is a counterfactual goal hypothesis. Assuming a full observation sequence, the explanation is given as a list of an observed action and its WoE value. Given that list, the explanation selection is based on the type of question to be answered.

\subsection{Explanation Generation}
We generate explanations by extending the WoE framework presented by \citeauthor{melis2021human} \shortcite{melis2021human}. By generating explanation lists using WoE, we can measure the relative importance of each observation to the goal hypotheses. This approach enables us to explain using the \textit{observational marker}, which is the predominant concept used in the explanation of the agent's goals (refer to Section \ref{freq}).

Referring to Equation~\ref{eq_woe}, we substitute the hypotheses $h$ and $h'$ with a predicted goal and counterfactual goal hypotheses, $g$ and $g'$. The evidence $e$ is replaced by an observed action $o_i$ and the posterior probabilities are represented as $P(g \mid O)$ and $P(g' \mid O)$. A complete explanation is defined as follows:

\begin{definition} A \emph{complete explanan} for a goal $g$ is a list of pairs $(woe(g/g' : o_i \mid O), o_{i})$, in which the conditional $woe(g/g' : o_i \mid O)$ for each paired hypothesis $g$ and $g'$ is computed for each added observation $o_i$ to the observed sequence $O$. The WoE is computed as follows:

\begin{equation}
woe(g/g' : o_i \mid O) = \log \frac{P(g \mid O_{i})}{P(g' \mid O_{i})}  
\end{equation}

\end{definition}

\begin{algorithm}[t]
\caption{Explanation Generation Algorithm}
\label{alg:algorithm}

\textbf{Input}: $O$, $G_p$, $G_c$, and $Posterior$ $probability$ over $G$\\
\textbf{Output}: Explanation list $\Omega$ of all $G_p$ paired with $G_c$
\begin{algorithmic}[1] 
\FOR{$o_i \in O$}
\STATE $\Omega \leftarrow \lbrack \rbrack$
\FOR{\textbf{each} \texttt{$g \in G_p$}}
 \FOR{\textbf{each} \texttt{$g' \in G_c$}}
\STATE \texttt{$\omega_{i} \leftarrow woe(g / g' : o_i \mid O)$} \\
\STATE \texttt{$\Omega \leftarrow \lbrack (g, g') = \langle \omega_{i}, o_{i} \rangle \rbrack$} 
  \ENDFOR
 \ENDFOR
\ENDFOR
\STATE \textbf{return} $\Omega$
\end{algorithmic}
\end{algorithm}

Informally, this defines a complete explanan for a goal $g$ as the complete list of computed WoE scores for each observation. An algorithm for extracting this is shown below  (Algorithm~\ref{alg:algorithm}).

In the navigational GR scenario presented previously (Figure \ref{figure7}), the WoE would be the same for all goal hypotheses after the first three observations, $o_{0}$ to $o_{3}$. This is because the Mirroring GR algorithm predicts them as equally likely since the first three observations are part of the optimal plan to achieve all three goals.
However, this uniformity does not hold for the rest of the observation sequence. For the observations $o_{4}$ to $o_{6}$, the Mirroring GR outputs would be goal $g2$ and $g3$ as equally likely since the observed actions are consistent with the optimal actions needed to reach either goal, with the counterfactual goal being $g1$. Table~\ref{woetab} presents the posterior probabilities and WoE values associated with either $g_2$ or $g_3$ as the leading goal candidate. The model computes the WoE value of each observed action for the pair of the predicted and counterfactual goals.

\subsection{Explanation Selection}
Explaining the output of a GR algorithm in terms of the \textit{complete} observation sequence can be tedious or even impossible, especially in scenarios where the domain model contains hundreds of thousands of states and actions. XAI best practice deems that for explanations to be effective they should be selective, focusing on one or two possible causes instead of all possible causes for a decision or recommendation \cite{miller2019explanation}. In the context of GR explanations, we found that people pointed to the \textit{observational marker} and the \textit{counterfactual observational marker} when they answered `why' and `why not' questions (refer to Sections \ref{concepts}, and \ref{freq}). To this end, we focus on selecting explanations to answer the `Why g?' and `Why not g'?' questions.

\begin{table}[h]
\centering
\begin{tabular}{@{}cccccc@{}}
\toprule
$o_{i} \in O$ &
  \multicolumn{1}{c}{$g$} &
  \multicolumn{1}{c}{$g'$} &
  \multicolumn{1}{c}{$P(g \mid o_{i})$} &
  \multicolumn{1}{c}{$P(g' \mid o_{i})$} &
  \multicolumn{1}{c}{$woe(g/g' : o_{i})$} \\ \midrule
$o_{4}$ & $g_2$ & $g_1$ & 0.36 & 0.27 & 0.28 \\
        & $g_3$ & $g_1$ & 0.36 & 0.27 & 0.28 \\
$o_{5}$ & $g_2$ & $g_1$ & 0.38 & 0.23 & 0.51 \\
        & $g_3$ & $g_1$ & 0.38 & 0.23 & 0.51 \\
$o_{6}$ & $g_2$ & $g_1$ & 0.40 & 0.20 & 0.69 \\
        & $g_3$ & $g_1$ & 0.40 & 0.20 & 0.69 \\ \bottomrule
\end{tabular}
\caption{Posterior Probabilities and Weight of Evidence (WoE) for Predicted and Counterfactual Goals After Observations $o_4,o_5$ and $o_6$ in the Navigational GR Example Depicted in Figure ~\ref{figure7}}
\label{woetab}
\end{table}

\subsubsection{`Why' Questions}
Answering \textit{why goal $g$}? questions, such as \textit{Why was goal $g$ predicted as the most likely goal over all other alternatives}?, rely on identifying the most important observation(s) that support the achievement of that goal. We call such observations \emph{observational markers} (OMs).

\begin{definition}[Observational Marker]
Given a complete explanan of $g$, the \emph{observational markers} (\textit{OMs}) are the observed actions that have the highest WoE value:
\begin{equation}
OM = \arg\max_{o_{i} \in O} [(g, g') = \langle \omega_{i}, o_{i} \rangle]
\end{equation}
\end{definition}

It is generated for every possible alternative and in case of having multiple such actions, we select them all. Consider the navigational GR scenario presented in Figure \ref{figure7}. Let us answer the question \textit{Why $g_2$}?. From the \textit{complete explanan} of $g_2$, shown in Table ~\ref{woetab}:
\[
\begin{array}{lll}
 (g_2, g_1) = & [\langle0.28, o_5\rangle, \langle0.51, o_6\rangle, \langle0.69, o_7\rangle,\langle0.85, o_8\rangle]  \\ 
 (g_2, g_3) = & [\langle0.18, o_8\rangle]  \\ 
\end{array}
\]
    After ranking them from highest to lowest, we obtain $\langle0.85, o_8\rangle$ that has the highest value. This indicates that this observation is the $OM$, as in the observation that best explains the predicted goal hypothesis $G_p = \{g_2\}$ instead of the counterfactual goal hypotheses, $G_c = \{g_1, g_3\}$. Therefore the explanation would be \textit{Because the agent has moved up from cell 26 to cell 17}.

\subsubsection{`Why Not' Questions}

The question of \textit{why not $g'$}? relies on identifying the most important observation(s) related to $g'$, which are  called \emph{counterfactual observational markers}.

\begin{definition}
Given a complete explanan of $g'$, the \emph{counterfactual observational markers} (\textit{counterfactual OMs}) are the observation(s) that have the lowest WoE value:
\begin{equation}
\text{counterfactualOM} = \arg\min_{o_{i} \in O} [(g, g') = \langle \omega_{i}, o_{i} \rangle]
\end{equation}
\end{definition}

There may be multiple such observations, in which case we select them all. Consider the navigational GR scenario (Figure \ref{figure7}) and the question \textit{Why not $g_1$ and $g_3$}? From the \textit{complete explanation} of $g_1$ and $g_3$, shown in Table ~\ref{woetab}:

\[
\begin{array}{lll}
 (g_2, g_1) = & [\langle0.28, o_5\rangle, \langle0.51, o_6\rangle, \langle0.69, o_7\rangle,\langle0.85, o_8\rangle]  \\ 
 (g_2, g_3) = & [\langle0.18, o_8\rangle]  \\ 
\end{array}
\]

After ranking them from lowest to highest, we obtain $\langle0.28, o_5\rangle$ as the lowest value for $g_1$ and $\langle0.18, o_8\rangle$ as the lowest value for $g_3$. This indicates that these observations are the \textit{counterfactualOM}, the observations that best explain the counterfactual goal hypotheses, $G_c = {g_1, g_3}$. Therefore, the explanation would be: \textit{Because the agent moved right from cell 23 to cell 24, away from $g_1$, and up from cell 26 to cell 17, away from $g_3$}.

\paragraph{Counterfactual Action}
Pointing to the lowest WoE action is not enough to answer \textit{why not} $g'$. Part of answering `why not' questions is the ability to reason about the \textit{counterfactual plan} that should have occurred instead of \textit{counterfactual OM} for $g'$ to be the predicted goal (refer to Section \ref{concepts}).

Building on this idea, we obtain the counterfactual action that should have happened instead of the observed one by planning the agent's route to $g'$ and simply taking the first action. We approach this problem by generating a plan for $g1$ from the state that precedes obtaining the \textit{counterfactual OM}, the state from which the lowest WoE is measured. We define the counterfactual action as follows.

\begin{definition}
Given a \textit{counterfacual OM} $o_i$ at state $s_{t-1}$ for counterfactual goal $g'$, a \emph{counterfactual action} $a'_{t}$ is the first action from the plan $\pi = \langle a'_{t}, a'_{t+1}, . . , g' \rangle$ that is generated by solving the planning problem $\langle \Xi, s_{t-1}, g'\rangle$, where $\Xi$ is the planning domain.
\end{definition}

Consider again the example from Figure~\ref{figure7}. The counterfactual action to $g_{1}$ would be the \textit{move up} action from cell 23 to 14, and to $g_{3}$ would be the \textit{move right} action from cell 26 to 27  (as indicated by the red arrows). Verbally, the answer to "Why not goal $g_{1}$?" would be: \emph{Because the agent moved right from cell 23 to cell 24. It would have moved up from cell 23 to 14 if the goal was $g_1$}, and to "Why not goal $g_{3}$?" would be: \emph{Because the agent moved up from cell 26 to cell 17. It would have moved right from cell 26 to 27 if the goal was $g_3$}

As noted in the navigational example (Figure \ref{figure7}), the counterfactual action for $g_3$ which is moving right from 26 to 27 (represented as a red arrow) is part of a suboptimal plan to $g_2$ (moving right to 27, up to 18, up to 9, and left to $g_2$). The framework operates by identifying the lowest observed evidence for a goal $g$ against goal $g'$ at time step $t$, and generating the counterfactual action from that point towards $g'$, even if that action is part of a plan for $g$.

\section{Evaluation}
In this section, we present a comprehensive analysis of the XGR framework as obtained through a combination of a computational study and three user studies to assess the effectiveness of our model.

\subsection{Computational Evaluation}
We evaluate the computational cost of the XGR model over eight online GR benchmark domains \cite{vered2018towards} to determine whether the cost of our approach is suitable for real-time explainability. The benchmark domains vary in levels of complexity and size, including different numbers of observations and goal hypotheses. We measure the overall time taken to run the XGR model. As the explanation model uses an off-the-shelf planner for counterfactual planning, we also separate the planner's cost and the explanation generation and show its effect on overall model performance.

Table~\ref{tab1} presents the run time performance of the XGR model over the benchmark domains. The run times vary greatly depending on the complexity of the domain, ranging from an average of 0.15 seconds over the 15 problems in the relatively simple Kitchen domain, to 221.77 seconds over the 16 problems in the complex Zeno-Travel domain (column 1). Regardless of the run time, adding our explainability model to the GR approach is typically not expensive, adding an increase of between 0.2\%-45\% (column 3). However, most of this increase can be attributed to calling the planner to generate counterfactual explanations (column 4). We can see that between 70\%-99\% of the XGR model run time is spent on planning. The varying percentage increases between domains like Zeno-Travel and Kitchen emphasize the relationship between domain complexity and planning time: the higher the domain complexity, the greater the influence the planner has. This highlights the significant impact of planner selection on the model performance. Since our model is independent of the underlying GR model, it has the potential to scale effectively with the integration of more efficient planners, such as domain-specific planners.

\begin{table}[ht]
\small
\centering
\begin{tabular}{@{}rrrrr@{}}
\toprule
\textit{\begin{tabular}[c]{@{}r@{}}Domain   \\      (\# problems)\end{tabular}} &
  \begin{tabular}[c]{@{}r@{}}Mirroring with \\      XGR (sec)\end{tabular} &
  \begin{tabular}[c]{@{}r@{}}XGR \\only (sec) \end{tabular} &
  \begin{tabular}[c]{@{}r@{}}Time \\ Increase (\%)\end{tabular} &
  \begin{tabular}[c]{@{}r@{}}Counterfactual \\    Planning (\%)\end{tabular} \\ \midrule
\begin{tabular}[c]{@{}r@{}}Campus\\       (15)\end{tabular}      & 0.21 (0.08)   & 0.019 (0.017)  & 10.15 & 87.11 \\
\begin{tabular}[c]{@{}r@{}}Ferry \\      (24)\end{tabular}       & 71.22 (36.16)  & 6.276 (8.070) & 09.66 & 99.69 \\
\begin{tabular}[c]{@{}r@{}}Intrusion \\      (45)\end{tabular}   & 0.69 (0.36)  & 0.215 (0.087) & 44.61 & 70.18 \\
\begin{tabular}[c]{@{}r@{}}Kitchen \\      (15)\end{tabular}     & 0.14 (0.07)   & 0.014 (0.002) & 11.12 & 73.61 \\
\begin{tabular}[c]{@{}r@{}}Rovers \\      (20)\end{tabular}      & 135.23 (73.11)  & 3.710 (7.271) & 02.82 & 99.64 \\
\begin{tabular}[c]{@{}r@{}}Satellite \\      (27)\end{tabular}   & 16.76 (10.05)  & 1.794 (10.052) & 11.98 & 99.27 \\
\begin{tabular}[c]{@{}r@{}}Miconic \\      (20)\end{tabular}     & 109.12 (22.61)  & 1.636 (2.861) & 01.52 & 98.72 \\
\begin{tabular}[c]{@{}r@{}}Zeno-Travel \\      (16)\end{tabular} & 221.77 (68.85) & 8.856 (11.721) & 04.15  & 99.65 \\ \bottomrule
\end{tabular}
\caption{Performance results of the XGR model for the \textit{Mirroring} Goal Recognition algorithm across eight benchmark domains. Column 1: Average and standard deviation of runtime with XGR and the mirroring GR algorithm. Column 2: Average and standard deviation of runtime with XGR only. Column 3: Percentage increase in runtime with XGR added to the GR approach. Column 4: Percentage of this additional runtime spent on counterfactual planning.}
\label{tab1}

\end{table}


\subsection{Empirical Evaluation}
We consider human studies experiments essential to the XGR model evaluation and conduct three human studies. The studies were conducted after obtaining institutional HREC approval.

\subsection{Study 1 - Generating Human-Like Explanations} \label{study1}

Our first study evaluates whether the model output is grounded on human-like explanations.

\subsubsection{Methodology and Experiment Design}

We used the method of annotator agreement and ground truth, where human annotations of representative features provided the ground truth for quantitative evaluation of explanation quality \cite{mohseni2020humangroundedevaluationbenchmarklocal}.

\paragraph{Task Setup} In this study, participants interacted with the output of the Mirroring GR algorithm across a series of problems within the Sokoban game domain (refer to Section \ref{sokoban} for additional details). The game was divided into three versions:
\begin{itemize}
    \item Game Version 1: Required the delivery of a single box to a single destination.
    \item Game Versions 2 and 3: Involved delivering two boxes to two sequential destinations, with interleaved plans to achieve each goal. The key distinction between these versions was that, in version 3, the agent could push multiple boxes, whereas in version 2, the agent could only push one box at a time.
\end{itemize}

This progression shifted the task from a straightforward navigation challenge to a more strategic one, where the player needed to manage multiple goals while striving to minimize the number of steps taken. Each game version included five scenarios of varying complexity, for a total of 15 scenarios. Each scenario presented a different goal recognition problem, with multiple competing goal hypotheses. Participants were required to answer "why" and "why not" questions regarding the predicted and counterfactual goal sets.

\begin{figure}[ht]
\centering
\includegraphics[width=0.75\textwidth]{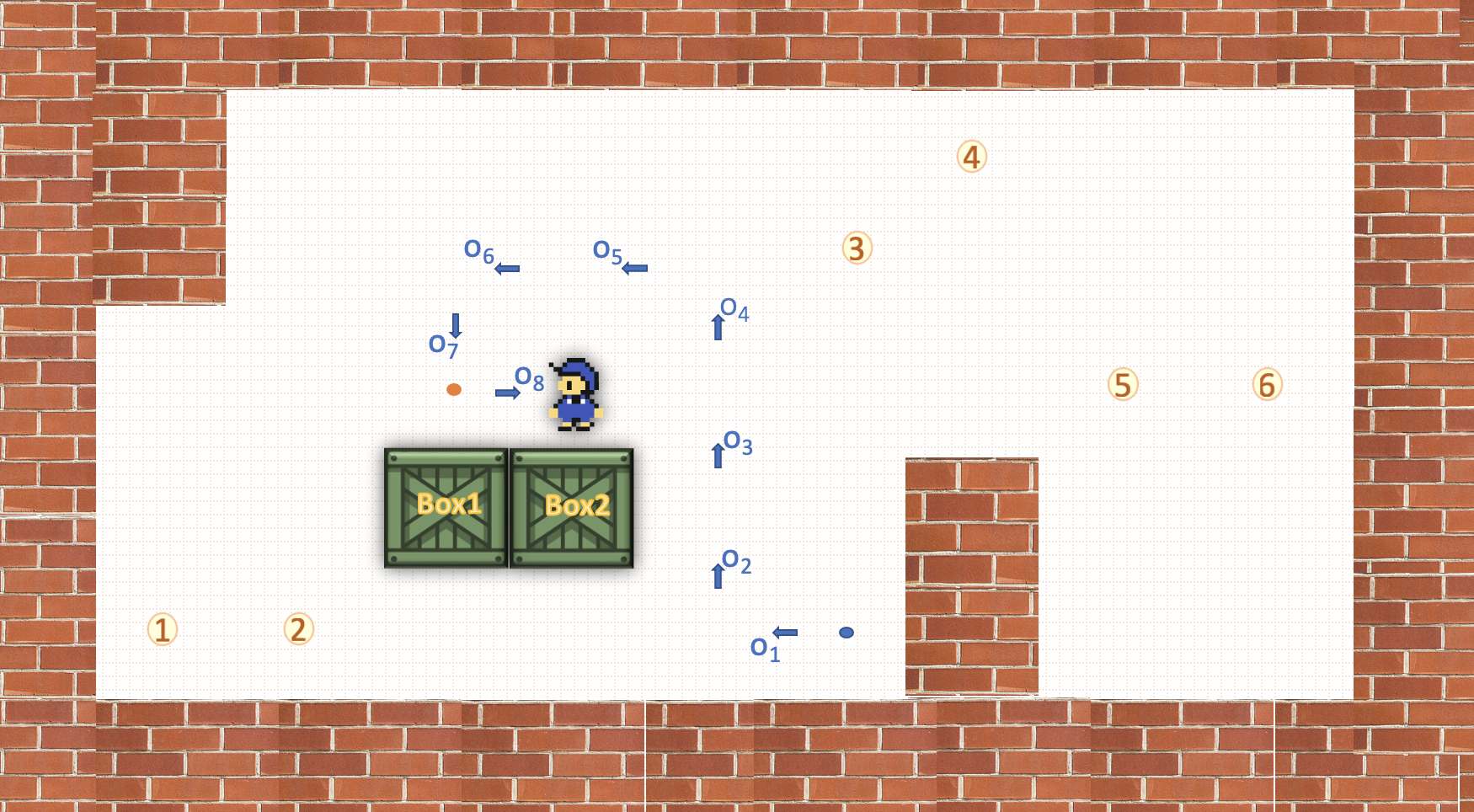} 
\caption{Sokoban game, scenario 5 (Game 3). The blue dot marks the initial state of the agent, and the orange dot marks the initial state of the box1 before pushing it down. There are 3 possible goals $((g_1,g_2),(g_3,g_4),(g_5,g_6))$. Blue arrows and hollow dots represent observations of the agent's actions.}
\label{figure8}
\end{figure}

\paragraph{Procedure} Each experiment lasted approximately 60 minutes and included the following four stages:
\begin{enumerate}
\item The game instructions were introduced to the annotators, along with a training scenario to help them understand the task.
\item The annotators watched a partial scenario (video clip) in which a Sokoban player attempted to achieve a goal (see Figure \ref{figure8}). The goals involved either delivering/pushing a box to a single destination cell or delivering/pushing two boxes to two different destination cells.
\item After watching the incomplete observation sequence, annotators were given the set of predicted and counterfactual goals generated by the Mirroring goal recognition algorithm.
\item The annotators were asked to identify the most important action, or optionally the two most important actions, from the observation sequence that addressed the questions: 'Why goal $g$?' and 'Why not goal $g'$?' Here, $g$ was the predicted goal, and $g'$ was the counterfactual goal. Additionally, participants were asked to annotate a counterfactual action for 'Why not goal $g$?'. This involved proposing a \textit{non-observed} action that they believed would indicate a move towards the alternative goal $g'$ (see Appendix C1. for an example scenario screenshot).
\end{enumerate}

\paragraph{Participants} We recruited three annotators (one male, two female) from the graduate student cohort at our university. Participants were aged between 29 and 40, with a mean age of 33. No prior knowledge of the task was required.

\paragraph{Metrics}  Using a majority vote, we combined participants' annotations into a single ground truth. Disagreements between annotations were typically resolved by an extra annotator. Then we calculated the mean absolute error (MAE) to assess how closely the obtained explanation of the XGR model matched the ground truth, defined by the agreement between annotators. The MAE was calculated as the average difference between each ground truth value ($a_{groundTruth}$) and the corresponding XGR value ($a_{XGR}$) over the length of the observation sequence ($n$). These values will be discussed in the following section.

\begin{equation}
MAE = \frac{1}{n}\sum_{i=1}^{n}\mid a_{groundTruth} - a_{XGR}\mid
\end{equation}

For evaluating the selection of a counterfactual action, since there is only one counterfactual action per plan, this was obtained by a binary agreement between the ground truth and the XGR model. For each domain, we calculated the percentage of agreements using the method described by \cite{araujo1985calculating}:
\begin{equation}
CF (\%) = \frac{agreements}{agreements + disagreements}\times 100%
\end{equation}

\subsubsection{Results}

We applied our XGR model to the online Mirroring implementation for each of the 15 scenarios. To determine the value of ($a_{XGR}$), we obtained the ranked explanation list from the XGR model based on the Weight of Evidence (WoE) values. For the question `Why goal $g$?', we ranked the explanations in descending order of their WoE values, assigning a rank of 1 to the explanation with the highest WoE (Observational Marker, OM). Similarly, for the question "Why not goal $g'$?", we ranked the explanations in ascending order of their WoE values, assigning a rank of 1 to the explanation with the lowest WoE (counterfactual OM). Using ground truth obtained through human annotation, we matched these ranks with their equivalents in the ranked explanation list for each question to determine $a_{groundTruth}$. We then calculated the Mean Absolute Error (MAE) for each question and across all 15 scenarios.

\begin{table}[]
\centering
\begin{tabular}{rrr}
\toprule
$o_{i}$ & WhyQ & WhyNotQ \\ \hline
$o_{8}$ & 1 & 6 \\
$o_{7}$ & 2 & 5 \\
$o_{6}$ & 3 & 4 \\
$o_{5}$ & 4 & 4 \\
$o_{4}$ & 5 & 3 \\
$o_{3}$ & 6 & 2 \\
$o_{2}$ & 7 & 1 \\ 
$o_{1}$ & 0 & 0 \\ 
\bottomrule
\end{tabular}
\caption{An explanation list generated by the XGR model which ranks each observation to answer `Why' and `Why not' questions for the example scenario depicted in Figure \ref{figure8}.}
\label{explanan}
\end{table}

\begin{example}
Consider the example in Figure \ref{figure8}. We obtained the ranked explanation list for both questions from our model (Table \ref{explanan}). The annotated actions from the ground truth are $o_{2}$ and $o_{7}$, which explain `Why $g_{1}$ AND $g_{2}$?', and $o_{2}$, which explains `Why not ($g_{3}$ AND $g_{4}$) OR ($g_{5}$ AND $g_{6}$)?'. We then determined ($a_{groundTruth}$), which is the equivalent rank of the annotated action in the list. For the "Why" question, these values are $o_{2} = 7$ and $o_{7} = 2$, and for the "Why not" question, the value is $o_{2} = 1$.
\end{example}

The Mean Absolute Error (MAE) is the average of the errors; hence, the larger the number, the larger the error. An error of 0 indicates full agreement between the models.

The results of the comparison are presented in Table ~\ref{tab:humanStudyResults}. Each row shows the MAE calculated for each game scenario. The `Why' and `Why not' columns represent the MAE for our model compared to the human ground truth, while the CF(\%) column represents the percentage of counterfactual action explanations that agree with the human ground truth.

For the majority of instances, the XGR model agreed with the ground truth obtained through human annotation. When answering `Why $g$?' questions, the model had a full agreement with the ground truth in 11 out of the 15 scenarios (73.3\%). For `Why not $g'$?' questions, the model had a full agreement with the ground truth in 14 out of the 15 scenarios (93.3\%). By `full agreement', we mean that the human annotators identify the same two actions as the primary explanation.

The CF column represents the percentage of counterfactual action explanations that agree with the human ground truth. Higher values are better, with 100\% indicating full agreement with the ground truth counterfactual actions. The model achieved full agreement in 11 out of the 15 scenarios (73.3\%), demonstrating excellent performance.

\begin{table}[ht]
\centering
\small
\begin{tabular}{lcccr} 
\toprule
\textbf{Game} & \textbf{Scenario} & ~\textbf{Why}~~ & ~\textbf{Why Not}~ & \textbf{CF} (\%)  \\ 
\midrule
1       & S1        & 0.00            & 0.00              & 100                                                                  \\
        & S2        & 0.00            & 0.00              & 100                                                                  \\
        & S3        & \textbf{0.37}            & 0.00              & 100                                                                  \\
        & S5        & \textbf{0.25}            & 0.00              & 100                                                                  \\
        & S5        & 0.00            & 0.00              & 100                                                                  \\[1mm]
2       & S1        & 0.00            & 0.00              & 66.6                                                                \\
        & S2        & 0.00            & \textbf{0.12}              & 33.3                                                                \\
        & S3        & 0.00            & 0.00              & 100                                                                  \\
        & S4        & 0.00            & 0.00              & 100                                                                  \\
        & S5        & 0.00            & 0.00              & 33.3                                                                \\[1mm]
3       & S1        & \textbf{0.50}            & 0.00              & 100                                                                  \\
        & S2        & 0.00            & 0.00              & 50                                                                   \\
        & S3        & 0.00            & 0.00              & 100                                                                  \\
        & S4        & 0.00            & 0.00              & 100                                                                  \\
        & S5        & \textbf{0.44}            & 0.00              & 100                                                                  \\ 
\midrule
Mean   &           & 0.10            & 0.008             & 89.40                                                               \\
SD     &           & 0.65            & 0.031             & 00.25                                                               \\
\bottomrule
\end{tabular}
\caption{The \textit{Why} and \textit{Why not} columns represent the mean absolute error (MAE) for XGR compared to the ground truth. The CF column represents the counterfactual action explanations percentage that agreed with the ground truth.}

\label{tab:humanStudyResults}
\end{table}

To better understand the performance of our model we investigated scenario 5 in game 3, which has a relatively high Mean Absolute Error (MAE) for the \textit{Why goal $g$?} question, we refer to Figure~\ref{figure8}. In this scenario, the agent delivers 2 boxes to 2 different locations and can push 2 boxes simultaneously. The blue arrows in the figure represent the observation sequence, indicating that the agent started at the blue circle and followed the arrows to its current location.

In this scenario, the most likely goal candidate, as predicted by the Mirroring GR algorithm, was delivering Box1 to $g_1$ and Box2 to $g_2$, i.e., $G_p = {(g_1,g_2)}$. The counterfactual goal candidates involved delivering the boxes to either $g_3$ and $g_4$ or $g_5$ and $g_6$, with $G_c = {(g_3,g_4), (g_5,g_6)}$. It is important to note that to push both boxes simultaneously to goal $(g_1,g_2)$, the agent would need to stand to the right of the boxes. Conversely, to push both boxes simultaneously to either $(g_3,g_4)$ or $(g_5,g_6)$, the agent would need to position itself to the left of the boxes.

The XGR model's explanation for the \textit{Why goal $g$?} question is highlighted by the green, hollow circle in the figure. This observation suggests that the agent aims to position itself to the right of both boxes, confirming the hypothesis that the goal is $(g_1,g_2)$. Considering the agent's ability to push multiple boxes, this observation constitutes the Observation Model (OM) with the highest Weight of Evidence (WoE).

On the other hand, the annotators established the ground truth explanation by choosing the second observation, $o_{2}$, for both the \textit{Why goal $g$}? and \textit{Why not $g'$}? questions. According to our model, this observation is the one with the lowest Weight of Evidence (WoE), actually making it the \textit{counterfactual observation} and the answer to the question \textit{Why not $g'$}? This is because this observation moves away from both goals $(g_3, g_4)$ and $(g_5, g_6)$.

Participants choosing to use the same answer for both \textit{Why goal $g$}? and \textit{Why not $g'$}? questions can also be found in other instances of discrepancies between the output of our model and the ground truth. The difference in explanations can be attributed to some confusion and/or preference between \textit{why}? and \textit{why not}? questions on the part of the participants.

To address this, we conducted a follow-up experiment where we presented the participants with the scenarios they were confused with (Table \ref{tab:humanStudyResults}, scenarios with bold values). For each scenario, we provided them with explanations from two systems: the first system's explanation from our model, and the second system's explanation from the ground truth. We then asked them which system provided a better explanation. All three participants preferred the explanations provided by our model which leads us to our model's explainability as perceived by users.


\subsection{Study 2 - Perceived Explainability}
The second human subject experiment aimed to evaluate the explainability of our model. We considered the following two hypotheses for our evaluation: 1) Our model (XGR) leads to a better understanding of a GR agent; and 2) A better understanding of a GR agent fosters user trust.

\subsubsection{Experiment Design and Methodology}

\paragraph{Conditions} We conducted a between-subjects study in which participants were randomly assigned to one of two conditions: 1) the explanation model (XGR), where an explanation was provided for the GR output; 2) the No Explanation model, where no explanation was provided for the GR output. We did not include a baseline for another explanation method due to the lack of existing alternatives.

\paragraph{Task Setup}
We used the Sokoban game as our test bed, following the same task setup as in our previous study (refer to Section \ref{study1}). Participants engaged with the output of the Mirroring GR algorithm across a series of problems within the Sokoban game domain. For each problem, we used a STRIPS-like discrete planner to generate ground truth plan hypotheses based on the domain theory and observations. Participants were then tasked with predicting the player's possible goal based on the observed behavior. Following this, they rated their trust on a 5-point Likert scale \cite{Hoffman2018MetricsFE}. In the XGR condition, participants also used a 5-point Likert scale to rate the given explanation according to their satisfaction with it \cite{Hoffman2018MetricsFE}.

\begin{figure}[ht]
\centering
\includegraphics[width=0.75\textwidth]{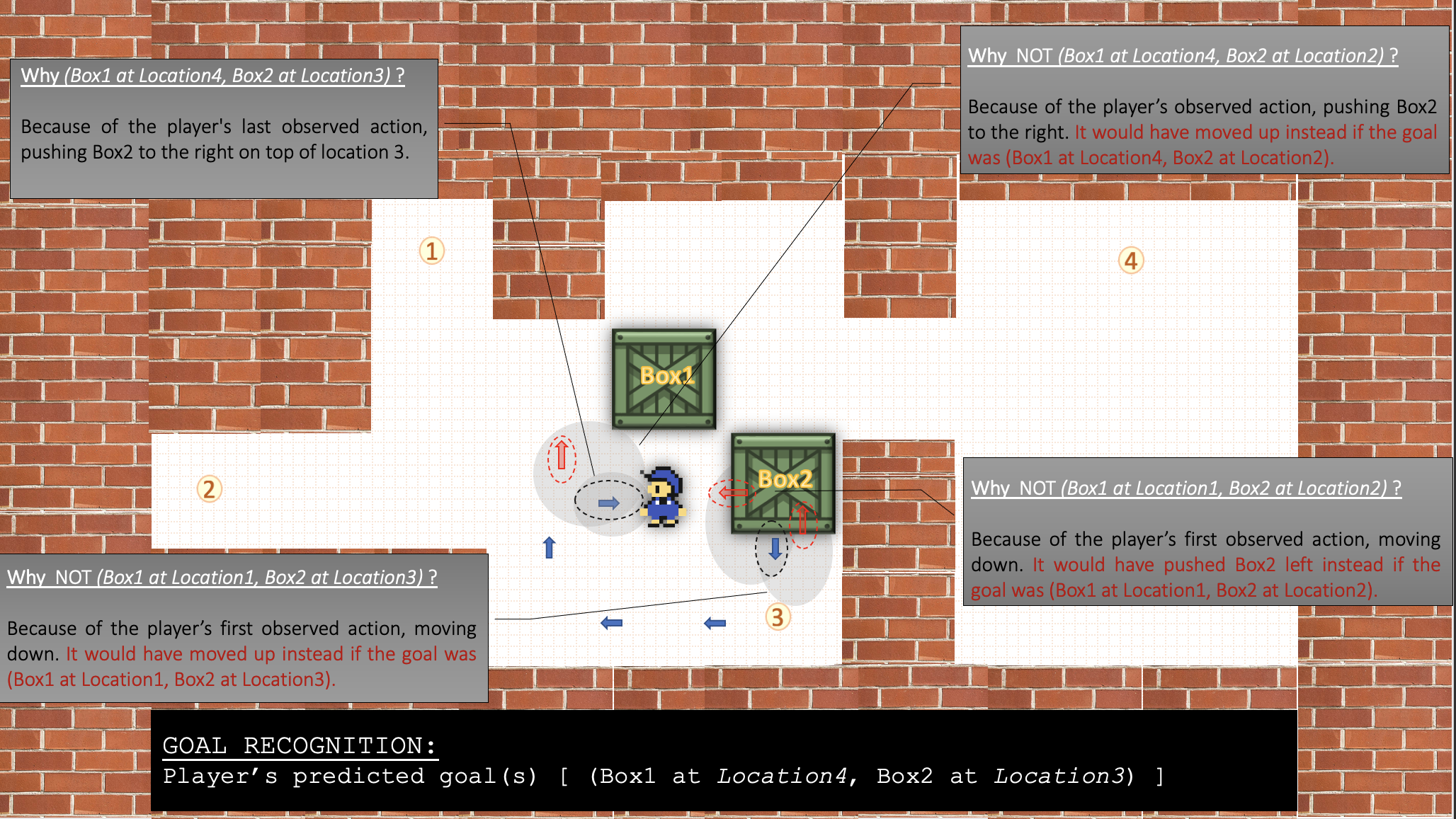} 
\caption{Sokoban game, scenario 5 (Game 3). The blue dot marks the initial state of the agent, and the orange dot marks the initial state of the box1 before pushing it down. There are 3 possible goals $((g_1,g_2),(g_3,g_4),(g_5,g_6))$. Blue arrows and hollow dots represent observations of the agent's actions.}
\label{figure}
\end{figure}

\paragraph{Procedure}
Participants were presented with six partial scenarios (video clips) showing a Sokoban player attempting to deliver boxes to designated locations. The experiment was divided into four phases:

\begin{enumerate}
\item \textbf{Phase 1:} Collection of demographic information and participant training. Using two video clips, the participant is trained to understand the player task and how to use GR and explainable system outputs. 
\item \textbf{Phase 2:} Presentation of a 10-second video clip of the Sokoban player's actions, along with the GR system output. Participants were asked to predict the agent's goals. For the No Explanation condition, participants made predictions without receiving any explanations. In the XGR condition, explanations for `why' and `why not' questions were presented (see Appendix C2. for example scenarios of the two conditions). Explanations were pre-generated by our algorithm and displayed on an annotated image of the video clip's final frame. The explanations were converted into natural language using a template, as exemplified in Figure \ref{figure}. 
\item \textbf{Phase 3:} Completion of the trust scale by participants.
\item \textbf{Phase 4 (XGR condition only):} Completion of the explanation satisfaction scale.
\end{enumerate}

\paragraph{Participants}
Prior to running the study, we performed a power analysis to determine the needed sample size. We calculated Cohen’s F and obtained an effect size of 0.35. Using a power of 0.80 and a significance alpha of 0.05, this resulted in a total sample size of 60 for the two conditions. We therefore recruited a total of 70 participants from Amazon MTurk, allocated randomly and evenly to each condition. To ensure data quality, we recruited only 'master class' workers whose first language is English and who have at least a 98\% approval rate on previous submissions. After excluding inattentive participants, we obtained 65 valid responses (No Explanation: 33, XGR: 32). Demographics included 28 males and 37 females, aged between 20 and 69, with a mean age of 40. Participants were compensated \$4.00 USD, with an additional bonus of \$0.20 USD for each correct prediction, up to a total of \$1.20 USD.

\paragraph{Metrics}
To test the first hypothesis, we used the task prediction method as described by \citeauthor{Hoffman2018MetricsFE} \citeyear{Hoffman2018MetricsFE}, which acts as a proxy measure for user understanding. Participants were scored one point for correct prediction and penalized one point for incorrect prediction. For the second hypothesis, we used the \textit{trust scale} from \citeauthor{Hoffman2018MetricsFE} \citeyear{Hoffman2018MetricsFE}, where participants rated their trust on a 5-point Likert scale ranging from 0 (Strongly Disagree) to 100 (Strongly Agree) based on four metrics. To evaluate the subjective quality of the explanations, participants used the \textit{Explanation Satisfaction Scale} from \citeauthor{Hoffman2018MetricsFE} \citeyear{Hoffman2018MetricsFE}, a 5-point Likert scale ranging from 0 (Strongly Disagree) to 100 (Strongly Agree) based on four metrics.

\paragraph{Analysis Method}
After conducting a homogeneity test to assess the variance of the collected data, we proceeded with Welch's t-test.


\begin{figure}[ht]
\centering
\includegraphics[width=0.6\textwidth]{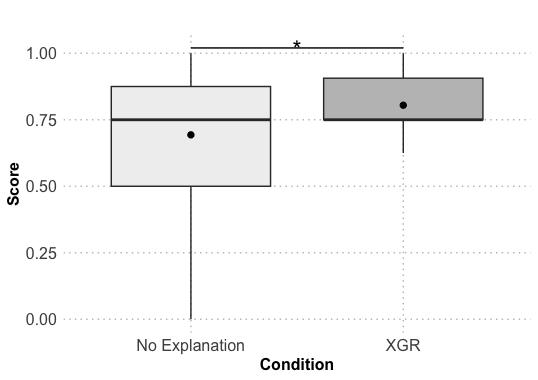} 
\caption{Task prediction scores for the two models (higher is better).}
\label{figure9}
\end{figure}

\subsubsection{Results}

\paragraph{Hypothesis 1: Our model (XGR) leads to a better understanding of a GR agent} 
Figure (\ref{figure9}) presents the variance in task prediction scores for the two models. A Welch's t-test indicated a significant difference (p-value $= 0.03$) in favor of the XGR cohort and the No Explanation cohort. These results suggest that our model (XGR) provides a significantly better understanding of the agent's behavior compared to the baseline model, as evidenced by the task prediction scores. Therefore, we accept our first hypothesis.

\begin{table}[ht]
\centering
\footnotesize
\begin{tabular}{@{}llll@{}}
\toprule
Understanding & Satisfying  & Sufficient Detail  & Complete    \\ \midrule
88.93 (10.8)  & 86.09 (12.2) & 87.93 (13.1) & 86.15 (19.6) \\ \bottomrule
\end{tabular}
\caption{Mean and standard deviation of explanation quality metrics for the XGR model}
\label{tab:eXpmetrics}
\end{table}

Table~\ref{tab:eXpmetrics} shows the mean and standard deviation of the explanation quality metrics for the XGR model on a Likert scale. Higher values indicate stronger agreement. These results suggest a satisfactory level across all four metrics.

\begin{figure}[ht]
    \centering
    \begin{subfigure}[b]{0.48\textwidth}
        \centering
        \includegraphics[width=\textwidth]{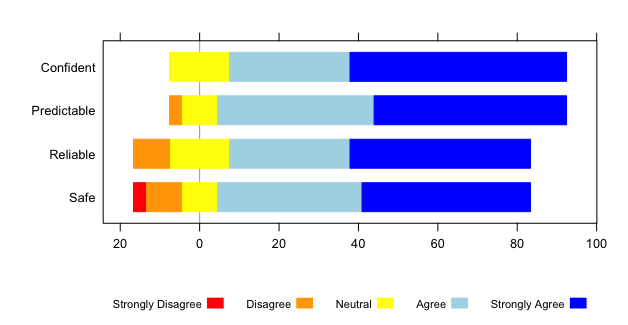}
        \caption{No Explanation Model}
    \end{subfigure}
    \hfill
    \begin{subfigure}[b]{0.48\textwidth}
        \centering
        \includegraphics[width=\textwidth]{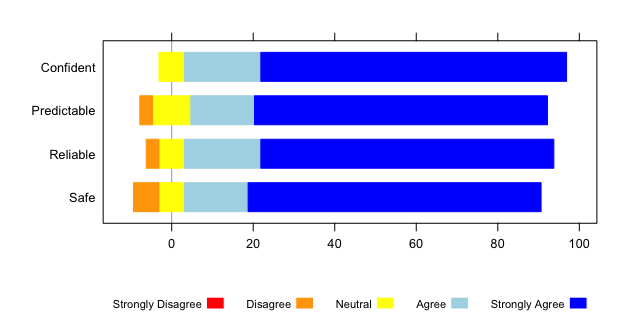}
        \caption{XGR Model}
    \end{subfigure}
    \caption{Likert scale of perceived trust metrics for the two conditions. The X-axis represents each Likert category's total counts of responses, adjusted to represent 0 as the midpoint.}
    \label{figure10}
\end{figure}

\paragraph{Hypothesis 2: A better understanding of a GR agent fosters user trust} \label{trust}
Figure (\ref{figure10}) illustrates the Likert scale data distribution of the participant's perceived level of trust across the two models. A Welch t-test yielded p-values (\(0.10, 0.18, 0.03, 0.02\)) for the trust metrics (\emph{Confident}, \emph{Predictable}, \emph{Reliable}, and \emph{Safe}) respectively. These results indicate a significant difference in the \emph{Reliable} and \emph{Safe}, and a marginally significant difference for the \emph{Confident} and \emph{Predictable} metrics. Thus, we also accept our second hypothesis.


\subsection{Study 3 - Effectiveness in Supporting Decision-Making} \label{study3}
This study aims to evaluate our model within the context of human-AI decision-making, focusing on several key hypotheses for empirical assessment. Specifically, we intend to investigate the impact of incorporating counterfactual explanations into our model. Our hypotheses are as follows: 1) Decision Accuracy: We hypothesize that our model enhances decision-making performance; 2) Decision Efficiency: We anticipate that the model will contribute to greater overall efficiency in task completion; 3) Appropriate Reliance on the GR Model: We propose that our model will promote more \textit{appropriate} reliance on the GR model; and 4) User Trust: We expect that our model will lead to increased trust in the GR agent. For this hypothesis, the aim is to further validate our findings (refer to Section \ref{trust}) in a different context. Additionally, we seek to address the research question: \textit{How do participants' reasoning processes vary across the four conditions?}

\paragraph{Conditions}
We conducted a between-subjects study where participants were randomly assigned to one of four conditions:
\begin{itemize}
    \item NoGR: Participants made their decisions with no goal recognition output received.
    \item GR: Participants made their decisions with goal recognition output received.
    \item XGR\_WoE: Participants made their decisions with goal recognition output received along with explanations of our XGR model based on WoE only.
    \item XGR\_WoE\_CF: Participants made their decisions with goal recognition output received along with explanations of our XGR model based on both WoE and counterfactual explanations.
\end{itemize}

We have the NoGR condition to understand the impact of AI assistance on decision-making processes. In addition, we decompose our model into XGR\_WoE, where the explanation is generated based on observational markers only, and XGR\_WoE\_CF, where explanations are generated based on both observational markers and counterfactuals, to facilitate comparison.

\begin{figure}[ht]
\centering
\includegraphics[width=0.9\textwidth]{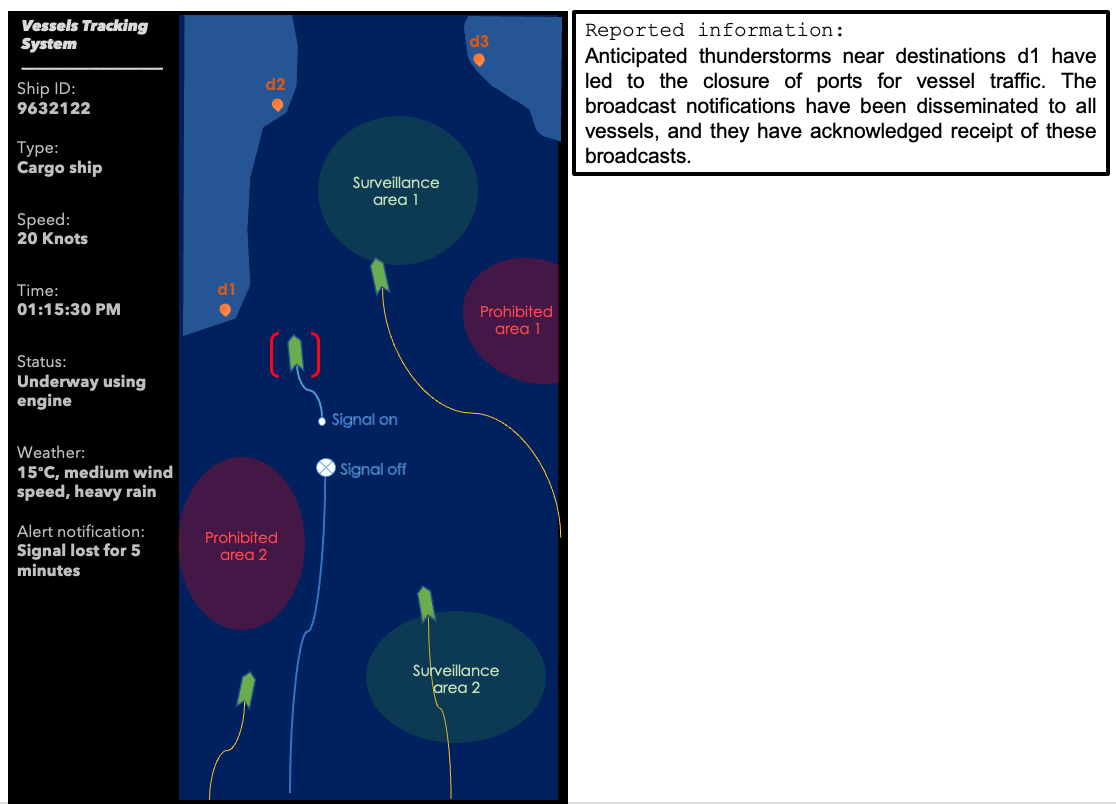} 
\caption{Maritime Domain, Scenario 2 (NoGR Condition).  }
\label{figureVessel}
\end{figure}

\paragraph{Task Setup}
This time we opted to test our model on a more complex domain,  inspired by real-world applications \cite{cordner2017maritime,Rosello2020}. We used a maritime domain as our test bed and designed eight scenarios to assess whether a vessel was involved in illegal maritime activities. In the eighth scenario, we introduced two detected vessels (8a, 8b), adding a level of complexity in dealing with multiple vessel interactions within the same environment. The scenarios include invading prohibited areas, deliberately avoiding surveillance areas, or concealing illegal operations by turning off signals. 

An example scenario is presented in Figure \ref{figureVessel}, which shows the detected vessel (red brackets) heading toward one of three destinations (port d1, d2, or d3). The GR task consists of 1) an initial state where the vessel is located at the start of its path on the map (depicted as a blue line); 2) a set of several distinct goal hypotheses, which include whether or not the vessel invades prohibited areas, whether or not the vessel is attempting to avoid surveillance areas, is the vessel attempting to conceal, and the destination it is heading to; and 3) the observed behavior sequence, which includes the vessel's path (represented as a blue line). The vessels navigate within a 2D environment, taking actions such as moving in four directions (up, down, left, right) and toggling their monitoring systems. Key model variables include the vessel's location, the locations of prohibited and surveillance areas, the destinations, and whether the vessel's signal has been lost or if it avoids certain areas. We used a STRIPS-like discrete planner to generate the ground truth plan hypotheses based on the domain theory and observations. Participants were tasked with assigning the likelihood of a detected vessel's possible destination based on its observed behavior, and the likelihood of the vessel being engaged in illegal activity and the need to be intercepted by the Coast Guard. Each participant was exposed to eight scenarios, which were randomly assigned and ordered to avoid order effects. We chose detecting illegal activity as our task since it involves complex, high-stakes scenarios requiring participants to judge based on limited and ambiguous information.

\paragraph{Procedure}
Participants were presented with eight scenarios of a vessel heading to one of the three destinations (seaports). The experiment consisted of four phases:

\begin{itemize}
\item \textbf{Phase 1:} Collection of demographic information and participant training. Participants were trained to understand the task using three training scenarios.
\item \textbf{Phase 2:} Participants were shown a static image simulating the tracking system display, along with the GR system output, referred to as the decision aid system in the second condition. Additionally, the explanation system output was included to answer "why" and "why not" questions in the third and fourth conditions. Participants were then asked to predict the vessel's goals regarding its destination and any potential illegal activities that might require interception by the Coast Guard. Each participant completed eight scenarios. Our algorithm pre-generated the explanations and presented them in natural language (refer to Appendix C3. for example scenarios of the four conditions).
\item \textbf{Phase 3:} In the last scenario, participants were given an open-ended question to justify their decision. 
\item \textbf{Phase 4:} Completion of the trust scale by participants for all conditions except the first (NoGR).
\item \textbf{Phase 5 (XGR conditions only):} Completion of the explanation satisfaction scale for explanation conditions (XGR\_WoE, XGR\_WoE\_CF).
\item \textbf{Phase 6:} After rating explanations, participants were asked two further questions: 
To what extent did you use the provided information? Options: not at all, minimally, moderately, substantially, extensively.
Please briefly describe how you incorporated the presented information into your decision-making process.
Participants were encouraged to provide detailed answers. We asked this question to gain insight into their thought processes and to assess their level of engagement with the task.
\end{itemize}

\paragraph{Participants}
We conducted a power analysis to determine the required sample size. Using Cohen’s F, we obtained an effect size of 0.20. With a power of 0.80 and a significance level of 0.05, we calculated a total sample size of 276 for the four conditions. We recruited 290 participants on Prolific, who were randomly and evenly allocated to each condition. To ensure data quality, participants had to reside in the US, UK, or Australia, be native English speakers, and have a minimum approval rate of 99\% with at least 1,000 previous submissions. After excluding inattentive participants, we obtained 280 valid responses (NoGR: 70, GR: 69, XGR\_WoE: 72, XGR\_WoE\_CF: 69). The demographic breakdown included 123 males, 154 females, and 3 self-specified, aged between 18 and 75, with a mean age of 37. Participants were compensated \$8.00 USD, with an additional performance-based bonus of up to \$3.00 USD.

\paragraph{Metrics}
We tested our first hypothesis using the Brier score function. The Brier score, which ranges from 0 (indicating worst performance) to 1 (indicating best performance), measures the accuracy of predictive probabilities for binary and multiclass outcomes. It is calculated as the mean squared distance between the predicted class probabilities and the actual outcomes (the ground truth) \cite{brier1950verification}:
\begin{equation}
    \text{Brier Score} = \sum_{i=1}^{c} (p_i - y_i)^2
\end{equation}

Where $c$ is the number of classes, $p$ is the predicted probability and $y$ is the ground-truth label represented as a vector $y = (y_1, \dots, y_c)$, where $y_i = 1$ for the true class and $0$ otherwise. We evaluated the quality of participants' decisions by measuring the Brier score of each decision across the eight scenarios.  A lower Brier score indicates better task performance. The Brier score assesses decision accuracy, rewarding a higher score for a correct answer with a high likelihood and imposing a lower penalty for an incorrect answer with a low likelihood. This reduces the impact of random guessing or uncertainty in responses. 

Outliers in response times can significantly skew statistical analyses, potentially leading to inaccurate interpretations of study results \cite{vankov2023hazards}. To address this, we assessed task efficiency in the second hypothesis by taking the logarithm of the time taken to complete each scenario.

To assess the appropriate reliance on the GR model for the third hypothesis, we used outcome-based measurement \cite{ma2024youVasconcelos}. This approach quantified human reliance on the system by categorizing it into overreliance (accepting the system's prediction when it is wrong) and underreliance (rejecting the system's prediction when it is correct).

\begin{equation}
    \text{Overreliance} = \frac{\text{Incorrect human decisions with incorrect GR predictions}}{\text{Total number of incorrect GR predictions}}
\end{equation}

\begin{equation}
    \text{Underreliance} = \frac{\text{Incorrect human decisions with correct GR predictions}}{\text{Total number of correct GR predictions}}
\end{equation}

We finally used the \textit{trust scale} from \citeauthor{Hoffman2018MetricsFE} \citeyear{Hoffman2018MetricsFE} to evaluate our fourth hypothesis. Based on four metrics, participants rated their trust on a 5-point Likert scale ranging from 0 (Strongly Disagree) to 100 (Strongly Agree). To evaluate the subjective quality of the explanations, participants used the \textit{Explanation Satisfaction Scale} from \citeauthor{Hoffman2018MetricsFE} \citeyear{Hoffman2018MetricsFE}, also a 5-point Likert scale ranging from 0 (Strongly Disagree) to 100 (Strongly Agree) based on four metrics.

\paragraph{Analysis Method}
Since the collected data did not meet the assumptions required for parametric testing, we used the non-parametric method's Kruskal-Wallis test. Pairwise differences were investigated using the Dunn test with Holm correction. Further, to assess the correlation between variables, we used  Pearson’s correlation coefficient.

\begin{figure}[ht]
\centering
\includegraphics[width=0.75\textwidth]{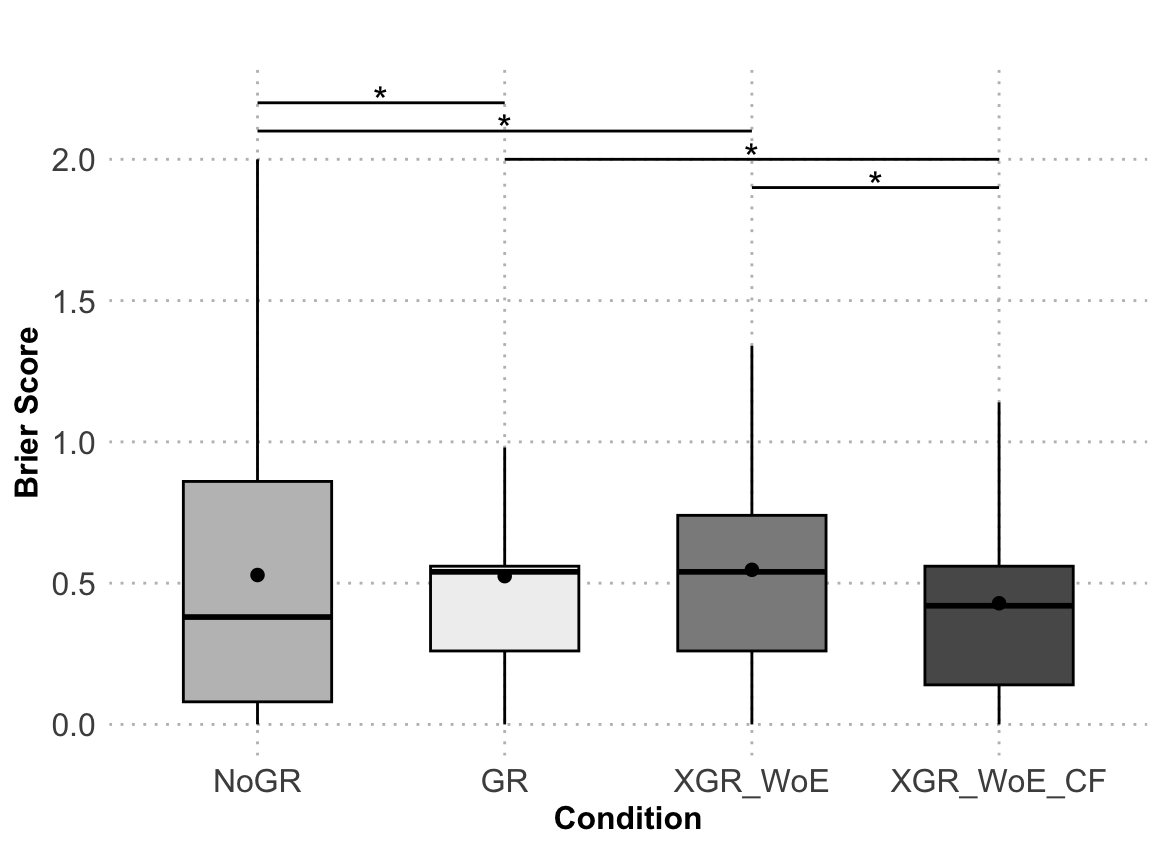} 
\caption{Brier score for the task of vessel destination detection across the four conditions, with means represented as dots (lower is better).}
\label{figure12}
\end{figure}

\subsection{Results}
In this section, we present the results of our experiments, addressing the four hypotheses and the research questions outlined in this study.

\subsubsection{Decision Accuracy}
We first discuss the results of the first hypothesis, where we investigate whether our model leads to a more accurate decision. 

\paragraph{The vessel destination}
The result of the statistical test for the vessel destination predictions indicates a significant difference between conditions (\(\chi^2 = 41.1\), \( p < 0.001\)). We then performed a pairwise comparison test and the results further indicate significant differences between conditions (Figure \ref{figure12}). There is a statistically significant difference between the XGR\_WoE\_CF cohort and the GR (with no explanations) cohort with \( p < 0.001 \). 
There is a marginally significant difference between the XGR\_WoE\_CF cohort and the NoGR cohort (no GR output and no explanations) with \( p  = 0.1 \). 
Thus, we accept our initial hypothesis that our model XGR\_WoE\_CF leads to more accurate decisions. 

Also, we observed a statistically significant difference between participant performance in the GR and NoGR cohorts with \( p = 0.002 \) indicating that participants performed worse with GR compared to NoGR. It has been argued that in some cases, people might perform worse with computer support even if the tool offers correct decisions. This can be due to various factors such as their skills and the computer's design \cite{alberdi2009people}. These results are reflected by measuring the appropriate reliance on GR model: while GR appears to help with underreliance, this benefit is countered by an increase in overreliance (Figure \ref{figure15}). This finding further strengthens previous findings \cite{vered2023effects} and suggests that while GR may reduce underreliance, it simultaneously leads to excessive trust in the GR outputs, ultimately resulting in poorer performance.

\begin{figure}[t]
\centering
\includegraphics[width=0.75\textwidth]{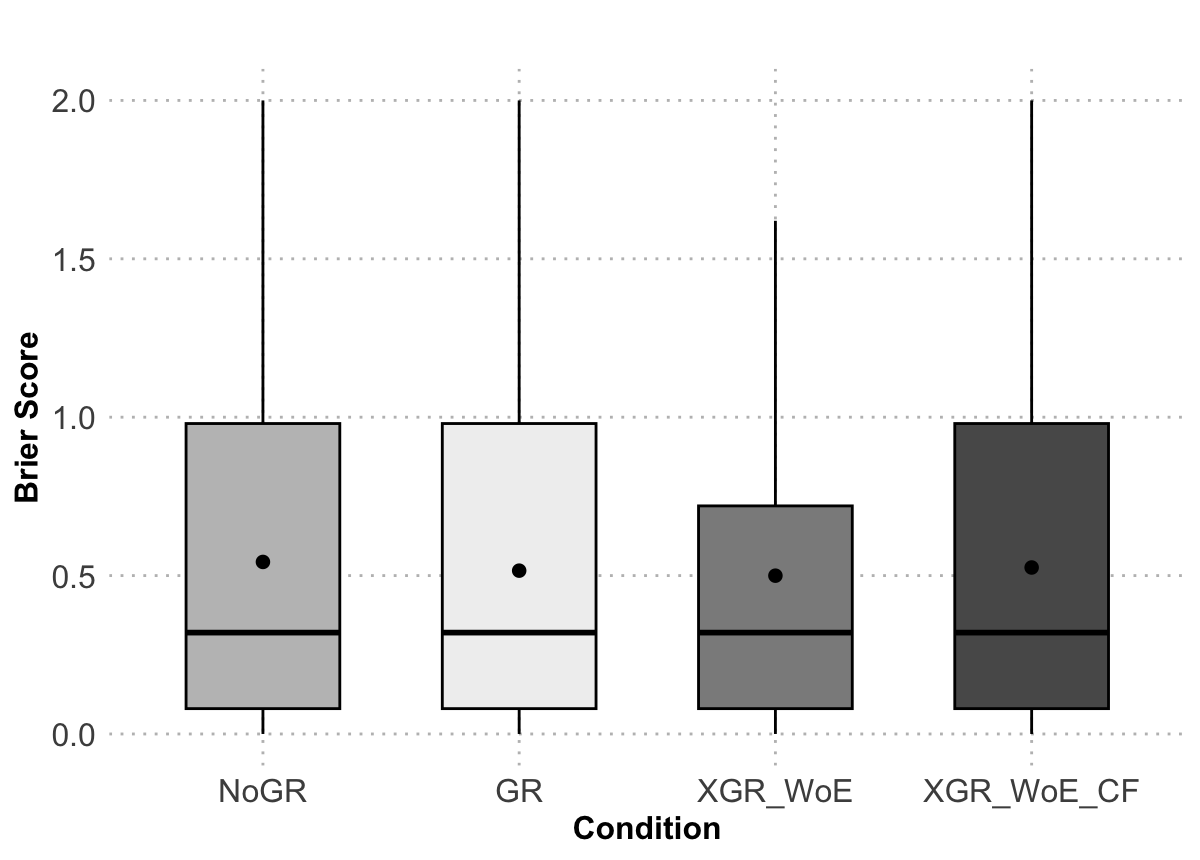} 
\caption{Brier score for the task of dispatching the coast guard,  across the four conditions with means represented as dots (lower is better).}
\label{figure13}
\end{figure}

\paragraph{Dispatching the Coast Guard}
When determining whether there is a need to dispatch the Coast Guard to intercept the vessel, the statistical analysis resulted in no statistically significant difference in the Brier scores among the different conditions (\( \chi^2 = 0.18 \), \( p = 0.98 \)). Figure \ref{figure13} illustrates these findings using a box plot. Consequently, we reject our hypothesis for this task that our model leads to more accurate decisions.

We hypothesize that the lack of significant difference may be attributed to the inherent complexity of the task, which involves processing multiple streams of information. Participants need to consider vessel behaviors such as invading prohibited areas, avoiding surveillance zones, and concealing activities. The cognitive load required to manage and synthesize all this information could lead to cognitive overload, potentially causing participants to rely less on analytical processing. This aligns with the notion that individuals are cost-sensitive decision-makers who weigh the effort required to use a decision aid system against its potential benefits \cite{kool2017cost}.

\begin{figure}[ht]
\centering
\includegraphics[width=0.75\textwidth]{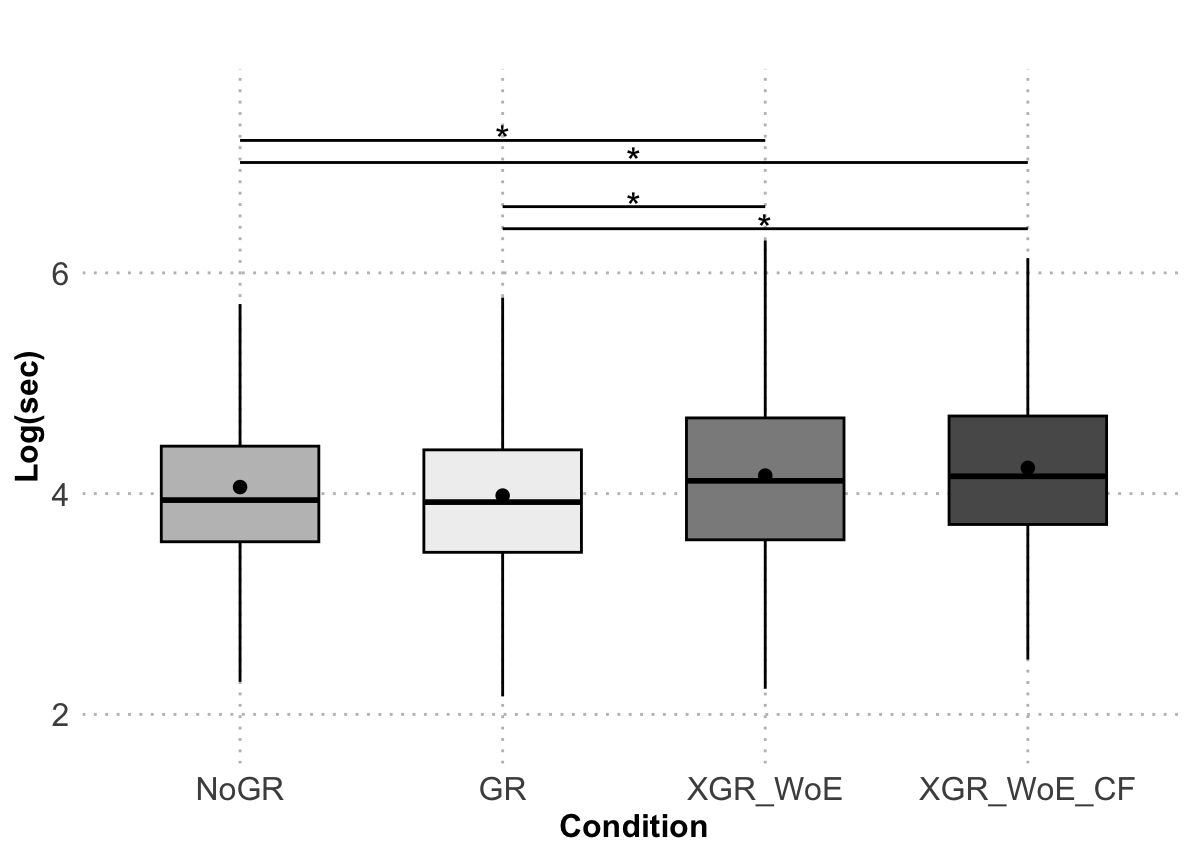} 
\caption{Completion time across the four conditions with means represented as dots (lower is better).}
\label{figure14}
\end{figure}

\subsubsection{Decision Efficiency}
For the second hypothesis, we evaluate how quickly participants made their decisions, hypothesizing that our model will contribute to greater overall efficiency in task completion. Efficiency was measured by the overall time spent on the tasks, recorded in seconds. 
statistical analysis revealed significant differences in completion time across the four conditions (\( \chi^2 = 39.28 \), \textit{p} \(< 0.001\)). Post hoc pairwise comparisons revealed that participants in the explanation cohorts, XGR\_WoE and XGR\_WoE\_CF, had significantly higher completion time compared to participants in the NoGR and GR cohorts (\( p < 0.001 \)). As shown in Figure \ref{figure14}, there are significant differences in completion time across the four conditions. However, no statistically significant difference was found between NoGR and GR (\( p = 0.13 \)), nor between XGR\_WoE and XGR\_WoE\_CF (\( p = 0.15 \)). Therefore, we reject the hypothesis that our models improve overall task efficiency.


\paragraph{Correlation of Decision Accuracy and Efficiency}
We further examined the relationship between Brier scores and task completion time to understand the influence of time spent on decision accuracy. Our findings suggest that longer task completion times were associated with lower Brier scores, indicating more accurate decisions. This is evidenced by a weak negative correlation between these variables. Notably, the correlation is statistically significant (\(p < 0.05\)) for certain conditions, as shown in Table \ref{tab:correlation} for the vessel destination task and Table \ref{tab:correlation2} for the Coast Guard task. These results highlight the importance of allocating sufficient time for information processing to improve decision-making accuracy.

\begin{table}[ht]
\small
\centering
\begin{tabular}{lcccc}
\toprule
Condition & Correlation & P-Value & Confidence Interval (95\%) \\
\midrule
NoGR & -0.0470 & 0.267 & [-0.129, 0.0360] \\
GR & \textbf{-0.100} & \textbf{0.0183} & \textbf{[-0.182, -0.0171]} \\
XGR\_WoE & \textbf{-0.179} & \textbf{0.0000158} & \textbf{[-0.257, -0.0986]} \\
XGR\_WoE\_CF & \textbf{-0.157} & \textbf{0.000222} & \textbf{[-0.237, -0.0741]} \\
\bottomrule
\end{tabular}
\caption{Pearson correlation coefficient between Brier score and completion time for vessel destination task. Significant values are indicated in bold.}
\label{tab:correlation}
\end{table}

\begin{table}[ht]
\small
\centering
\begin{tabular}{lcccc}
\toprule
Condition & Correlation & P-Value & Confidence Interval (95\%) \\
\midrule
NoGR & -0.0600 & 0.156 & [-0.142, 0.0230] \\
GR & \textbf{-0.129} & \textbf{0.00234} & \textbf{[-0.210, -0.0463]} \\
XGR\_WoE & -0.0609 & 0.144 & [-0.142, 0.0209] \\
XGR\_WoE\_CF & \textbf{-0.124} & \textbf{0.00363} & \textbf{[-0.205, -0.0406]} \\
\bottomrule
\end{tabular}
\caption{Pearson correlation coefficient between Brier score and completion time for Coast Guard task. Significant values are indicated in bold.}
\label{tab:correlation2}
\end{table}

\subsubsection{Appropriate Reliance on GR Model}
For the third hypothesis, we examine whether our model influences the appropriateness of human reliance, specifically by reducing overreliance and underreliance on the GR model. We measure overreliance as the fraction of incorrect human decisions that align with incorrect GR predictions, calculated over the total number of incorrect GR predictions. Conversely, underreliance is measured as the fraction of incorrect human decisions that occur despite correct GR predictions, calculated over the total number of correct GR predictions.

\begin{figure}[ht]
    \centering
    \begin{subfigure}{0.45\textwidth}
        \centering
        \includegraphics[width=\textwidth]{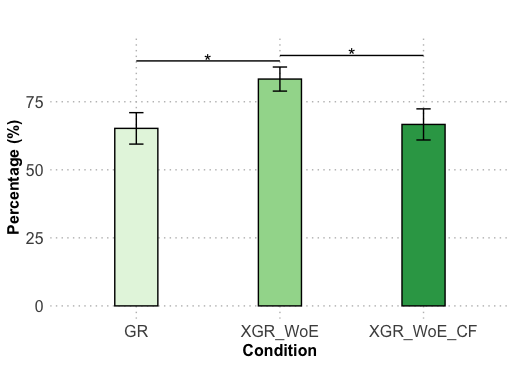}
        \caption{Overreliance Results}
    \end{subfigure}
    \hfill
    \begin{subfigure}{0.45\textwidth}
        \centering
        \includegraphics[width=\textwidth]{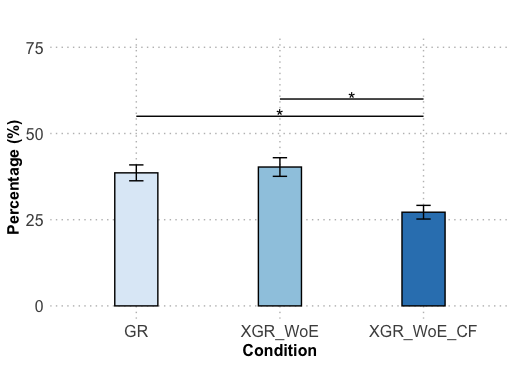}
        \caption{Underreliance Results}
    \end{subfigure}
\caption{Appropriatehuman reliance for the task of vessel destination prediction (lower is better). Error bars indicate standard errors of the mean.}
\label{figure15}
\end{figure}

\paragraph{Vessel destination prediction task. }
For the task of predicting the vessel's destination, the statistical analysis revealed significant differences between the conditions (\(\chi^2 = 7.08\), \(p = 0.03\)) in terms of  \textbf{overreliance}. Figure \ref{figure15}(a) shows the results of the pairwise comparisons, indicating significant differences between GR paired with XGR\_WoE (\(p = 0.02\)) and XGR\_WoE paired with XGR\_WoE\_CF (\(p = 0.03\)). Explanations increased the chance that humans would rely on system predictions \cite{bansal2021does}, but the counterfactual explanation in our model (XGR\_WoE\_CF) helped reduce that overreliance. Therefore, we accept our hypothesis only for XGR\_WoE\_CF.

The statistical analysis further revealed significant differences between the conditions (\(\chi^2 = 16.1\), \(p < 0.001\)) in terms of measuring \textbf{underreliance}. Figure \ref{figure15}(b) shows the results of the pairwise comparisons, indicating significant differences between GR paired with XGR\_WoE\_CF (\(p < 0.001\)) and XGR\_WoE paired with XGR\_WoE\_CF (\(p < 0.001\)). XGR\_WoE\_CF 
significantly reduces underreliance compared to GR 
and XGR\_WoE. 
We again accept our hypothesis only for XGR\_WoE\_CF.

\begin{figure}[ht]
    \centering
    \begin{subfigure}{0.45\textwidth}
        \centering
        \includegraphics[width=\textwidth]{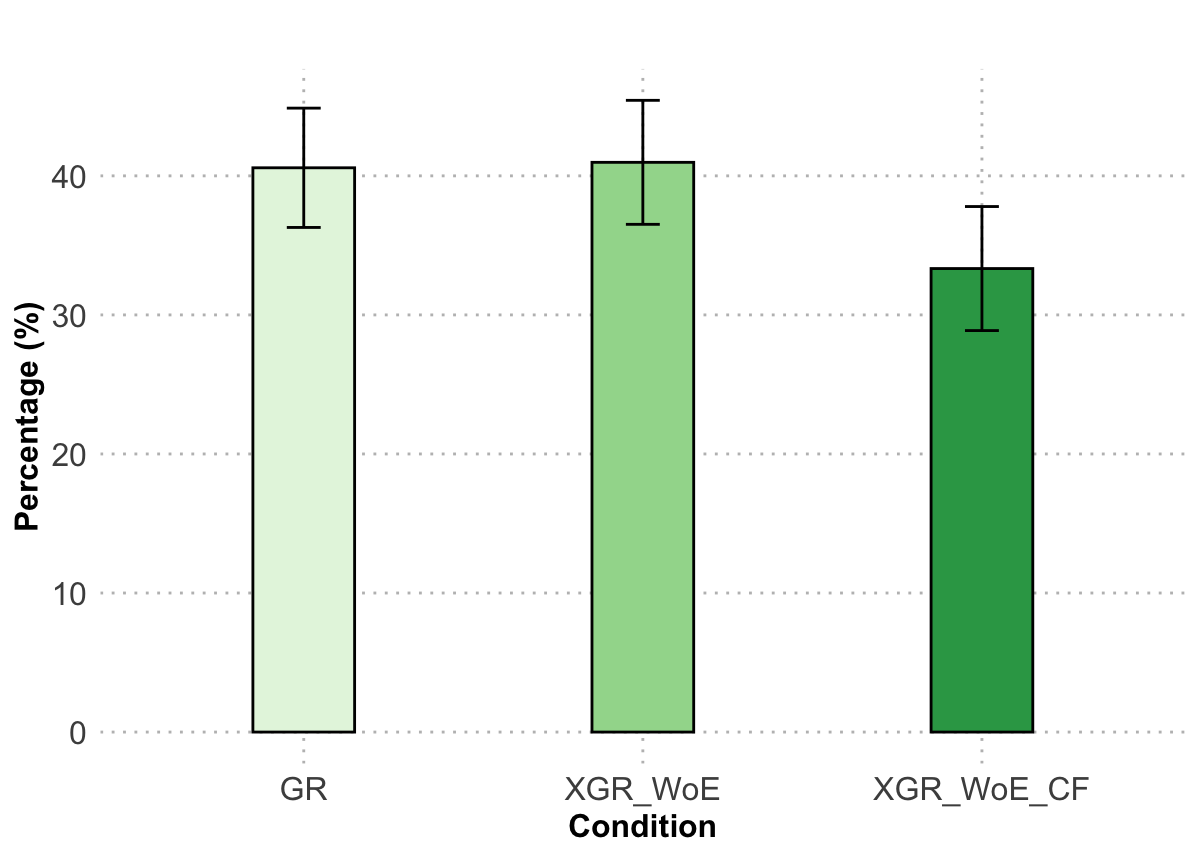}
        \caption{Overreliance Results}
    \end{subfigure}
    \hfill
    \begin{subfigure}{0.45\textwidth}
        \centering
        \includegraphics[width=\textwidth]{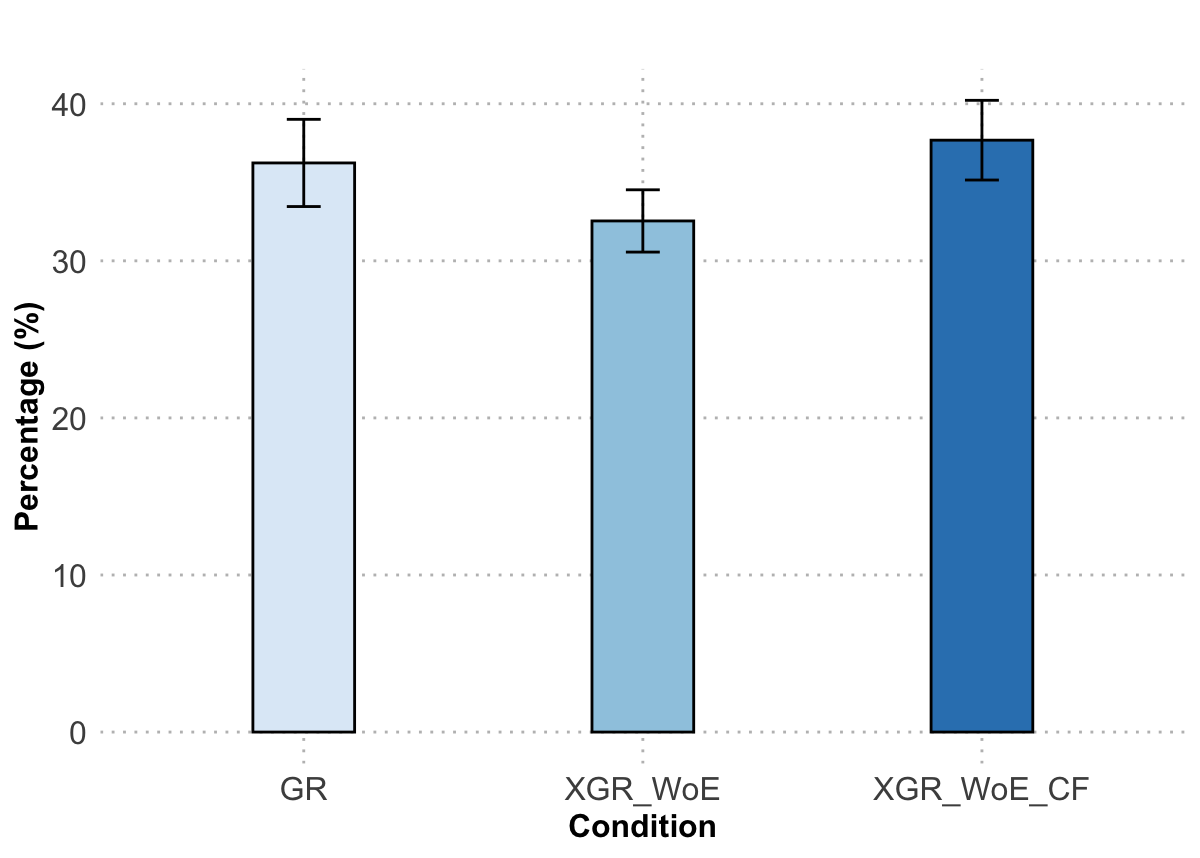}
        \caption{Underreliance Results}
    \end{subfigure}
\caption{Approriate human reliance for the task of dispatching the coast guard lower is better). Error bars indicate standard errors of the mean.}
\label{figure16}
\end{figure}

\paragraph{Dispatching the Coast Guard}

We also measured over and under reliance of users on the GR model for the task of deciding whether the Coast Guard should be dispatched to intercept the vessel a difference in the mean values between the three conditions, no statistically significant differences were found following the statistical analysis (\(\chi^2 = 2.12\), \(p = 0.35\)) to measure the \textbf{overreliance} (see Figure \ref{figure16} (a)). Thus, we reject our hypothesis.

The same result was found when measuring the effect of reducing \textbf{underreliance} (see Figure \ref{figure16} (b)). Despite different mean values between the three conditions, no statistically significant differences were found following the statistical analysis (\(\chi^2 = 1.38\), \(p = 0.50\)). We therefore reject our hypothesis of promoting appropriate reliance in this task.  While our model XGR\_WoE\_CF appears to help with overreliance, this effect is countered by the result of underreliance, which reflects the accuracy results we observed (Figure \ref{figure13}). When people find the task and explanation cognitively demanding, they tend to rely on the system's prediction without carefully verifying it \cite{vasconcelos2023explanations}.

\begin{figure}[ht]
    \centering
    \begin{subfigure}[b]{0.45\textwidth}
        \centering
        \includegraphics[width=\textwidth]{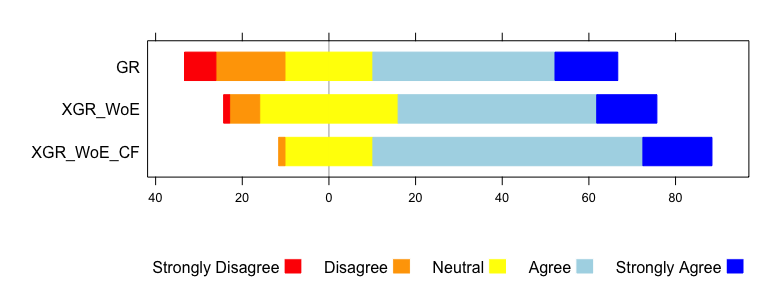}
        \caption{Confident}
        \label{fig:confident}
    \end{subfigure}
    \hfill
    \begin{subfigure}[b]{0.45\textwidth}
        \centering
        \includegraphics[width=\textwidth]{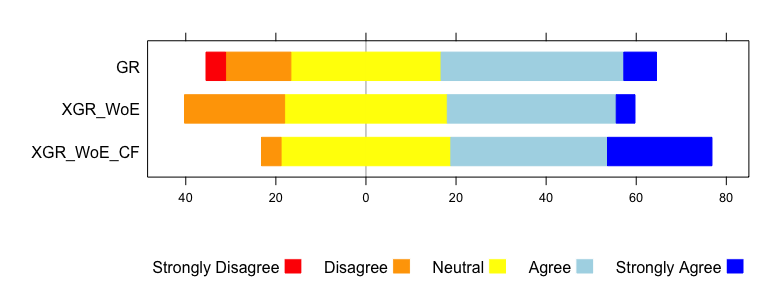}
        \caption{Predictable}
        \label{fig:predictable}
    \end{subfigure}
    
    \vspace{1em} 
    
    \begin{subfigure}[b]{0.45\textwidth}
        \centering
        \includegraphics[width=\textwidth]{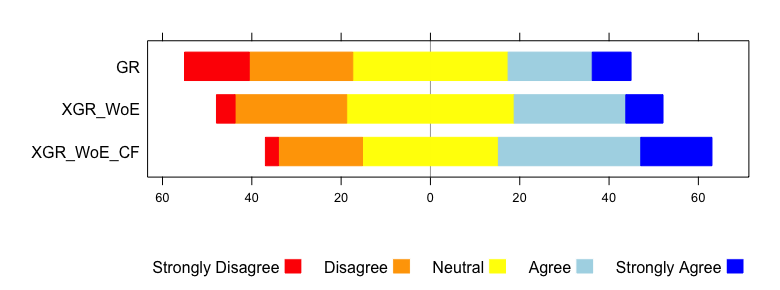}
        \caption{Reliable}
        \label{fig:reliable}
    \end{subfigure}
    \hfill
    \begin{subfigure}[b]{0.45\textwidth}
        \centering
        \includegraphics[width=\textwidth]{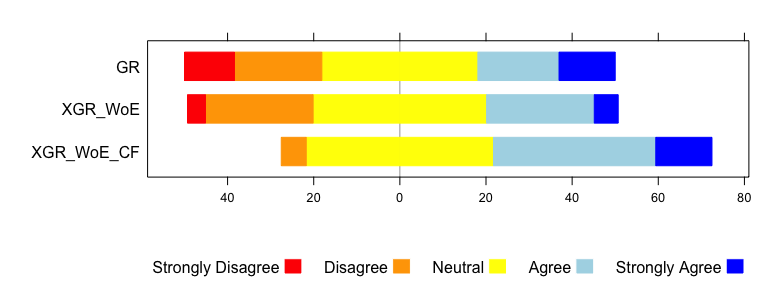}
        \caption{Safe}
        \label{fig:safe}
    \end{subfigure}
    \caption{Likert scale of perceived trust metrics across the three conditions. The X-axis represents each Likert category's total counts of responses, adjusted to have 0 as the midpoint.}
    \label{figure17}
\end{figure}

\subsubsection{User Trust}
We now report the results of our evaluation to determine whether our explanation model promotes trust in the GR agent. Figure \ref{figure17} illustrates the distribution of Likert scale data across conditions. The statistical test was performed, yielding a significant result ($p < 0.001$) for all metrics, indicating a significant difference among the three conditions for the Confident, Predictable, Reliable, and Safe metrics. Subsequently, we conducted pairwise comparison tests, and the results are summarized in Table \ref{tab:trustscale}. These findings highlight that the XGR\_WoE\_CF condition shows significant results across all metrics. Therefore, we accept the hypothesis that our model, XGR\_WoE\_CF, effectively promotes trust in the GR agent.

\begin{table}[h!]
\small
\centering
\caption{Pairwise comparisons with Post-hoc test for trust metrics}
\begin{tabular}{llccc}
\toprule
\textbf{Metric} & \textbf{Comparison} & \textbf{Z} & \textbf{P.adj} \\
\midrule
\multirow{3}{*}{\textbf{Confident}} 
 & GR vs XGR\_WoE     & 0.11  & 1.00          \\
 & GR vs XGR\_WoE\_CF & -3.06 & $<$\textbf{0.01} \\
 & XGR\_WoE vs XGR\_WoE\_CF & -3.20 & $<$\textbf{0.01} \\
\midrule
\multirow{3}{*}{\textbf{Predictable}} 
 & GR vs XGR\_WoE     & 0.94  & 1.00          \\
 & GR vs XGR\_WoE\_CF & -2.47 & \textbf{0.03}    \\
 & XGR\_WoE vs XGR\_WoE\_CF & -3.44 & $<$\textbf{0.01} \\
\midrule
\multirow{3}{*}{\textbf{Reliable}} 
 & GR vs XGR\_WoE     & -1.13 & 0.25          \\
 & GR vs XGR\_WoE\_CF & -3.22 & $<$\textbf{0.01} \\
 & XGR\_WoE vs XGR\_WoE\_CF & -2.12 & 0.06          \\
\midrule
\multirow{3}{*}{\textbf{Safe}} 
 & GR vs XGR\_WoE     & 0.60  & 0.55          \\
 & GR vs XGR\_WoE\_CF & -2.89 & $<$\textbf{0.01} \\
 & XGR\_WoE vs XGR\_WoE\_CF & -3.52 & $<$\textbf{0.01} \\
\bottomrule
\label{tab:trustscale}
\end{tabular}
\end{table}

\subsubsection{Explanation Satisfaction}

We now report the results of the self-reported metrics of explanation satisfaction. We performed statistical tests to examine whether there are any significant differences between the explanation models XGR\_WoE and XGR\_WoE\_CF for the explanation quality metrics: understanding, satisfying, sufficient detail, and completeness. The obtained p-values ($0.20, 0.31, 0.34, 0.24$) for these four metrics indicate no significant differences between the two models. Although the differences are not statistically significant, Table \ref{tab:eXpmetrics2} shows that the XGR\_WoE\_CF model is generally better perceived across these metrics, with more consistent responses from participants. This mirrors the results in decision accuracy, as participants feel they have a better understanding of the agent.

\begin{table}[ht]
\centering
\caption{Mean (SD) of explanation quality cross conditions}
\begin{tabular}{lcccc}
\toprule
\textbf{Condition} & \textbf{Complete} & \textbf{Satisfying} & \textbf{Sufficient Detail} & \textbf{Understand} \\ 
\midrule
\textbf{XGR\_WoE} & 64.2 (21.4) & 65.8 (21.8) & 67.3 (22.0) & 69.2 (19.8) \\
\textbf{XGR\_WoE\_CF} & 67.5 (19.7) & 69.6 (17.4) & 71.0 (18.4) & 73.6 (15.2) \\
\bottomrule
\end{tabular}
\label{tab:eXpmetrics2}
\end{table}

\begin{figure}[ht]
\centering
\includegraphics[width=0.6\textwidth]{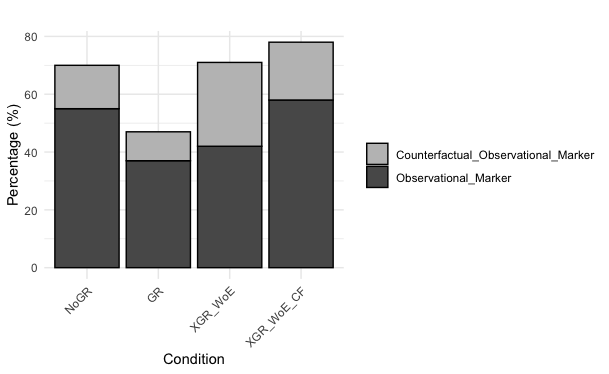} 
\caption{Frequency of using Observational Markers across conditions.}
\label{figure18}
\end{figure}

\subsubsection{Reasoning Variations Across Different Conditions (Research Question 1)} 
Participants answered an open-ended question about their decision justification in the last scenario. After excluding answers with fewer than three words or containing gibberish, a total of 244 textual data points were collected. We focused on understanding the reasoning process and explanation concepts used across the four conditions. We systematically coded the responses following a hybrid approach of deductive and inductive thematic analysis approach \cite{braun2006using}, labeling responses based on our pre-established main concepts (\textit{observational marker and counterfactual observational marker}). We then analyzed participants' responses and identified a few differences in their justifications across the four conditions. 

\paragraph{Explaining Using Observational Markers.}
We found that across conditions, participants in their justifications mainly relied on the most important evidence for their goal hypothesis (i.e., \textit{observational markers}) (see Figure \ref{figure18}). From the given information, they focused only on what they believed were the key features that increased the likelihood of their hypothesis. In the explanation conditions (XGR\_WoE and XGR\_WoE\_CF), they were also able to reason about the \textit{counterfactual observational marker} more frequently, which is the evidence against their hypothesized counterfactual goal. This suggests that the explanations, even without explicit counterfactuals (CF), prompted people to think in a counterfactual way, compared to just giving the goal. An example from the data corpus (NoGR condition): ``\textit{The speed change of vessel B along with signal loss needed investigating.}''.

\paragraph{GR Output Initiates Reasoning.}
Due to the presence of GR output in the GR, XGR\_WoE, and XGR\_WoE\_CF conditions, we observed that participants often began their justifications by referencing the GR output, explicitly stating their agreement or disagreement. This suggests that the GR output may serve as a reference point to initiate their reasoning. For example, in the GR condition, one participant stated, ``Because the intention to hide was above 50 percent, it would be best if the vessel was intercepted''. Similarly, in the XGR\_WoE\_CF condition, another participant noted, ``The decision aid (GR system) has made an error in suggesting there is a 100\% probability that they have invaded a prohibited area when they were avoiding a prohibited area. This vessel does not require interception''.

\paragraph{Explanations can help ensure that GR output is trustworthy with sufficient certainty.}
Some participants made efforts to ensure that GR output could be trusted by explicitly discussing their perceptions of the reliability of the GR output to decide how best to use it. They used the provided information to guide their trust in GR. This was particularly evident in conditions with sufficient confidence in XGR\_WoE and XGR\_WoE\_CF. An example from data corpus (XGR\_WoE\_CF condition): ``Vessel B's signal is lost for 30 minutes near a prohibited area, high probability of intentional concealment, also supported by the decision aid (GR system), and hence Coast Guard should be alerted''.

\subsection{Discussion}
Our three studies demonstrated the validity of our XGR model. We found that the model enhanced transparency and understanding of GR agents. Additionally, we found that the counterfactual explanation was a crucial component of our model when addressing "why not" questions, as evidenced in the third study (see results of XGR\_WoE\_CF, Section \ref{study3}). Counterfactual explanations were preferred by users since they offered coherent and extensive explanations that encompassed multiple instances, aligning with human preferences for comprehensive insights \cite{MILLER20191}.

However, our findings in the context of decision-making support, the task of dispatching the Coast Guard, were inconclusive. This could arise from a combination of factors, including task difficulty, participant engagement, or the possibility that the explanations provided were not sufficiently useful or aligned with the decision-making requirements of the task. The task complexity may have been beyond the participants' capabilities or understanding, leading to challenges in accurately making decisions. Difficult tasks can result in increased cognitive load and errors, causing people to struggle with accurately evaluating the AI model's output, potentially obscuring the model's true effectiveness \cite{vasconcelos2023explanations}. This suggests that the cost of verifying the AI model's output in this challenging task may have been so high that it outweighed the benefits, potentially making manual task completion more efficient. 

Furthermore, the complexity of the task may have influenced participant engagement. Participants who found the task too challenging might have been less engaged, impacting their performance. Engagement issues could also be linked to the task's perceived relevance and length. Low engagement can result in superficial responses and a lack of attention to detail, as evidenced by some participants completing the tasks very quickly. These observations indicate that both task difficulty and participant engagement played a role in the inconclusive results, suggesting that further investigation is necessary.

\section{Conclusion}
We developed an explainable model for GR agents that can generate explanations to answer why and why-not questions. The model is grounded in empirical data from two different human-agent studies. We evaluated our model computationally across eight online GR benchmark domains. Additionally, we conducted three human studies to investigate how the XGR model generates human-like explanations, increases user understanding and trust in GR agents, and improves the decision-making process.

While results indicate a significantly better performance of our model compared to the baselines, further research is needed to address the impact of task difficulty and participant engagement on decision-making support. This includes designing appropriately challenging yet understandable tasks and ensuring that participants are adequately motivated and engaged throughout the study. Additionally, we plan to extend our model to handle scenarios with partial observability, where some information may be missing or hidden. Another avenue for future work is to evaluate our model by incorporating additional concepts and examining how they affect user performance and satisfaction.


\newpage

\appendix
\section*{Appendix A. Human-Agent Study 1: Sokoban Game}

The Sokoban game is a single-player game where the player is responsible for pushing a box to one of the goal locations on the map, which are identified by numbers. Across fifteen Sokoban game scenarios (five for each game version), the participant's task is to predict which goal the player is pushing the box toward, based on the player's observed behavior.

Below is a screenshot of example scenarios across the three game versions and the corresponding participant task. All data will be made available upon request.

\begin{figure}[H]
    \centering
    \includegraphics[width=0.7\textwidth]{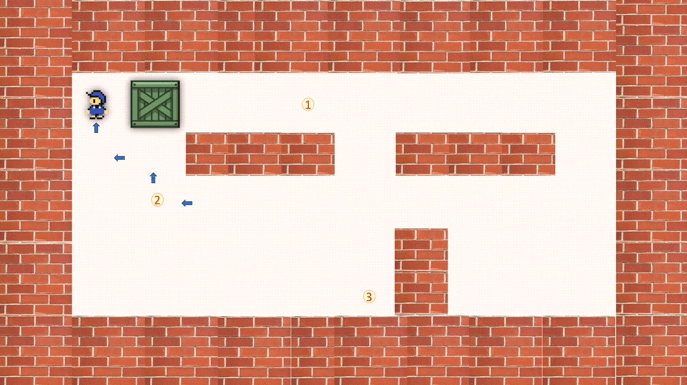} 
    \caption{Example scenario from Game Version 1: The player’s task is to move one box to the goal location.}
    \label{fig:appendix_image}
\end{figure}

\begin{figure}[H]
    \centering
    \includegraphics[width=0.7\textwidth]{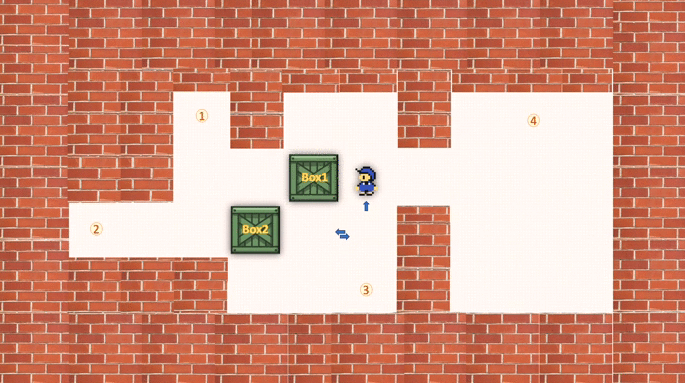} 
    \caption{Example scenario from Game Version 2: The player's task is to move two boxes to their designated goal locations, with the ability to push only one box at a time.}
    \label{fig:appendix_image1}
\end{figure}

\begin{figure}[H]
    \centering
    \includegraphics[width=0.7\textwidth]{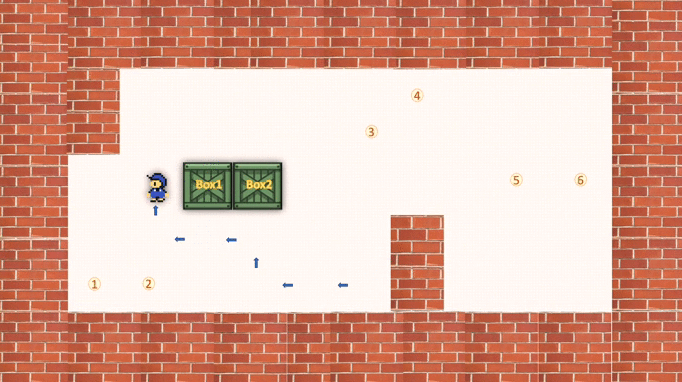} 
    \caption{Example scenario from Game Version 3: The player's task is to move two boxes to their designated goal locations, with the ability to push them simultaneously.}
    \label{fig:appendix_image2}
\end{figure}

\begin{figure}[H]
    \centering
    \includegraphics[width=0.7\textwidth]{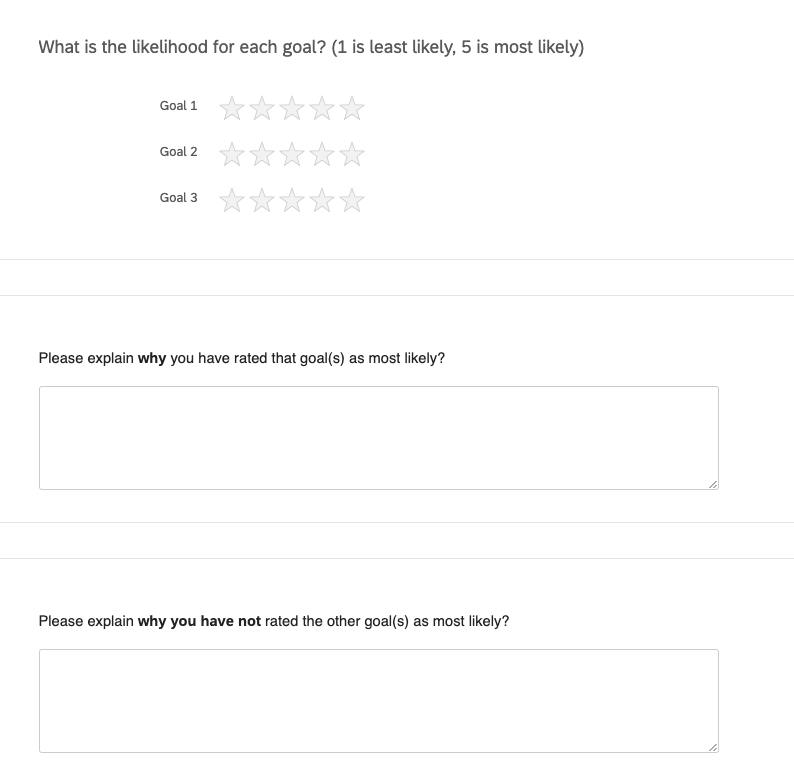} 
    \caption{Participant's task from the Dual Condition (Game Version 1): Assign likelihoods to goal locations and answer questions about why and why not.}
    \label{fig:appendix_image3}
\end{figure}


\section*{Appendix B. The Weight of Evidence (WoE): Formula Derivation}

The WoE is defined for some evidence $e$, a hypothesis $h$, and its logical complement $\overline{h}$ \cite{wod1985weight} as follows: 

\begin{equation}
woe(h : e) = \log \frac{Odds(h \mid e)}{Odds(h)}  
\end{equation}

\noindent
where the colon is read as ``provided by'', and $Odds(.)$ denotes the hypothesis odds:

\begin{equation}
Odds(h \mid e) = \frac{P(h \mid e)}{P(\overline{h} \mid e)}      \textrm{~~~(Posterior odds)}
\end{equation}

\begin{equation}
Odds(h) = \frac{P(h)}{P(\overline{h})}     \textrm{~~~~~(Prior odds)}
\end{equation}

This is the ratio of the posterior to the prior odds. The odds corresponding to a probability $p$ are defined as $p/(1-p)$ --- the probability of an event occurring divided by the probability of it not occurring.

Using Bayes' rule, $woe(h : e)$ can also be defined as:

\begin{equation}
woe(h : e) = \log \frac{P(e \mid h)}{P(e \mid \overline{h})}
\end{equation}

WoE can also contrast $h$ to an arbitrary alternative hypothesis $h'$ instead of its complement $\overline{h}$. Thus, we can talk generally about the strength of evidence in favor of $h$ and against $h'$ provided by $e$: 

\begin{equation}
woe(h/h' : e) = woe(h : e \mid h \vee h')
\end{equation}

The WoE generalizes to include cases when it can be conditioned on additional information $c$:

\begin{equation}
woe(h : e \mid c) = \log \frac{P(e \mid h, c)}{P(e \mid h', c)}
\end{equation}

From various properties of WoE, the following two properties are essential to our model:

\begin{equation}
\label{eq:woe-h-h-prime}
woe(h/h' : e) =  \log \frac{P(e \mid h)}{P(e \mid h')} 
\end{equation}


\begin{equation}
 \log \frac{P(h)}{P(h')} + \log \frac{P(e \mid h)}{P(e \mid h')} = \log \frac{P(h \mid e)}{P(h' \mid e)}
\end{equation}

By substituting using Equation~\ref{eq:woe-h-h-prime}, we get:

\begin{equation}
 \log \frac{P(h)}{P(h')} + woe(h/h' : e) = \log \frac{P(h \mid e)}{P(h' \mid e)}
\end{equation}

\begin{equation}
woe(h/h' : e) = \log \frac{P(h \mid e)}{P(h' \mid e)} - \log \frac{P(h)}{P(h')} 
\end{equation}

Using the log quotient property, we simplify the equation as follows: 

\begin{equation}
woe(h/h' : e) = \log \frac{\frac{P(h \mid e)}{P(h' \mid e)}}  {\frac{P(h)}{P(h')} }
\end{equation}

If we have uniform prior probabilities, we can simplify further and compute WoE for a pair of hypotheses (conditioned on $c$) as follows:

\begin{equation}
woe(h/h' : e \mid c) = \log \frac{P(h \mid e, c)}{P(h' \mid e, c)}  
\end{equation}



\section*{Appendix C. Empirical Evaluation}

In this section, we provide a brief description of the tasks and screenshots of selected scenarios from the three conducted studies. All data will be made available upon request.

\subsection*{C1. Study 1 - Generating Human-Like Explanations}
Fifteen Sokoban game scenarios (five for each game version) are presented for the goal recognition (GR) agent to determine the most likely goal based on the observed sequence of actions. Given the recognized goal set $G$ and the counterfactual goal set $G'$ at a specific time step, the participant's task involves answering the following questions for each scenario by annotating the map:\begin{itemize}
\item For the `why' question: \textit{Annotate} the most important action that justifies why the goal is considered the most likely, and the second most important action, if any.

\item For the `why not' question:
\begin{enumerate}
    \item \textit{Annotate} the most important action that justifies why the goal is considered less likely, and the second most important action, if any.
    \item Annotate the counterfactual action that should have occurred instead.
\end{enumerate}
\end{itemize}

Below is a screenshot of an example scenario from Game Version 1, along with the corresponding participant task and an example answer.

\begin{figure}[htbp]
    \centering
    \includegraphics[width=0.7\textwidth]{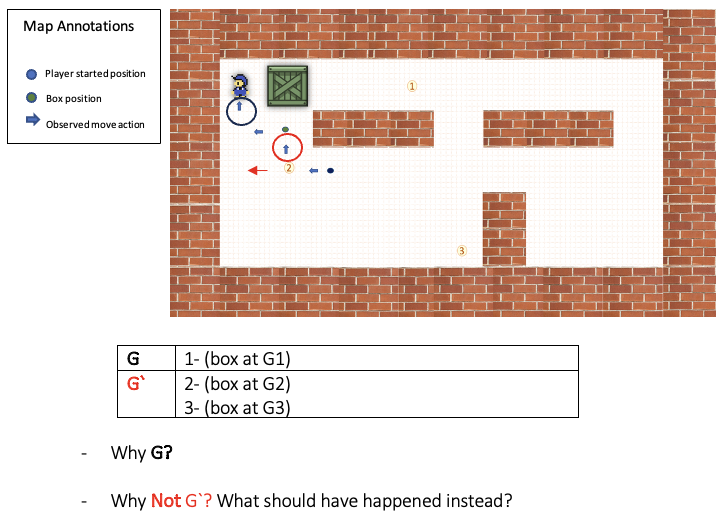} 
    \caption{Example scenario from Game Version 1. The black annotation addresses the question `Why $G$?' while the red annotation addresses the question `Why not $G'$?'}
    \label{fig:appendix_image4}
\end{figure}


\subsection*{C2. Study 2 - Perceived Explainability}
In this study, we provided six short gameplay videos of a Sokoban game (two scenarios for each game version). For the GR condition, we presented the output of the Goal Recognition system for each scenario. For the XGR condition, we provided both the GR output and an explanation of this output that addresses two questions: `Why?' and `Why not?'

Below is a screenshot showing an example scenario from both conditions, along with the corresponding participant task.

\begin{figure}[H]
    \centering
    \includegraphics[width=0.65\textwidth]{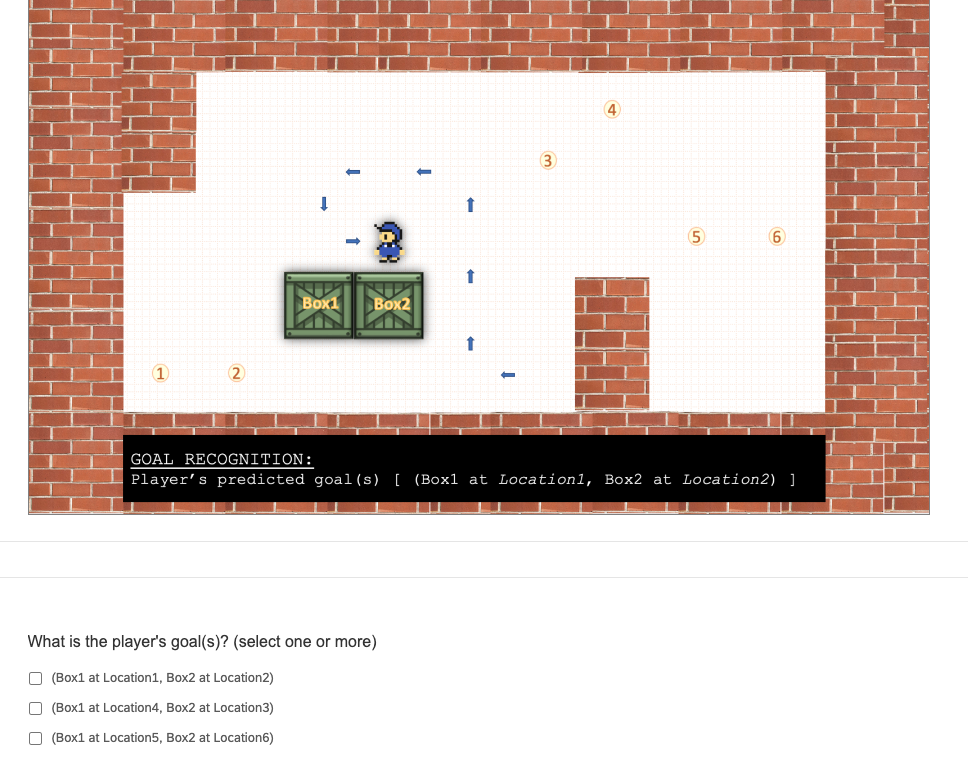} 
    \caption{Example scenario from Game Version 3 (GR Condition), where the participant's task is to predict the player’s goal locations.}
    \label{fig:appendix_image5}
\end{figure}

\begin{figure}[H]
    \centering
    \includegraphics[width=0.65\textwidth]{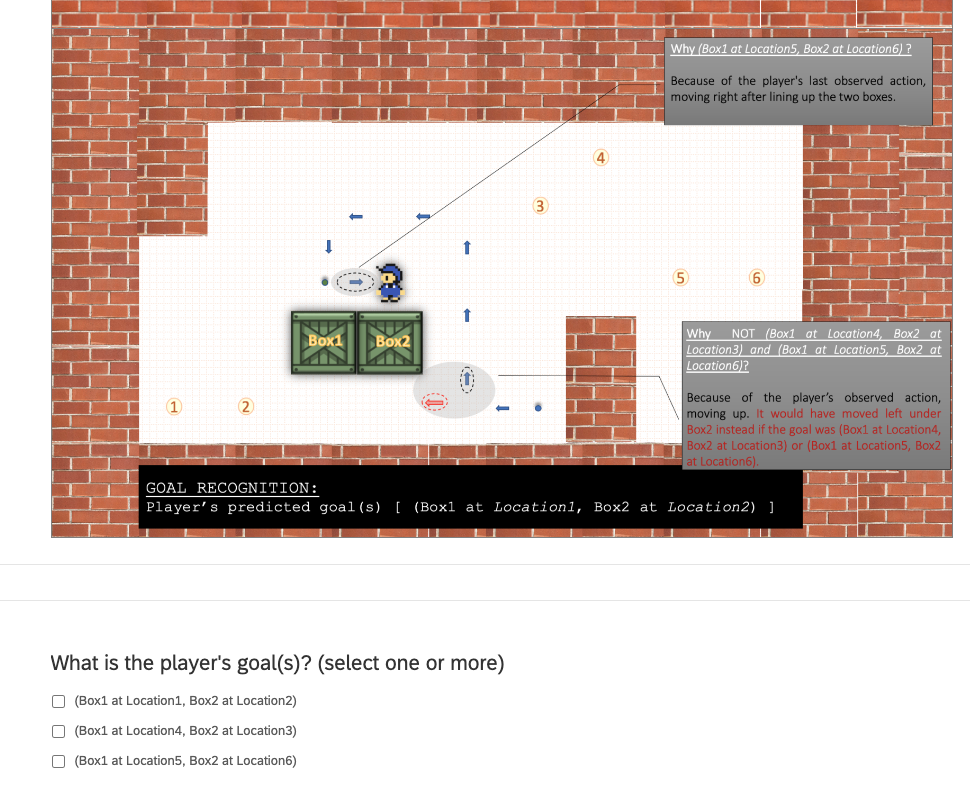} 
    \caption{Example scenario from Game Version 3 (XGR Condition), where the participant's task is to predict the player’s goal locations.}
    \label{fig:appendix_image6}
\end{figure}

\subsection*{C3. Study 3 - Effectiveness in Supporting Decision-Making} 

In maritime surveillance, control centers monitor vessels to detect illegal fishing activities. The main task is to decide on the interception of vessels engaging in such activities, which may involve deploying the Coast Guard. The ocean is divided into prohibited areas, where fishing is forbidden due to conservation concerns, and surveillance areas, where vessels must report their position and activities to authorities. Illegal vessels may either invade prohibited areas or avoid surveillance areas to avoid detection. Additional challenges include potential signal interruptions, which may be due to weak coverage or bad weather, rather than illegal activities.

The participant's task is to incorporate the evidence provided in the scenario to make a final decision regarding the vessel’s destination and the necessity of dispatching the Coast Guard. The likelihood of a vessel engaging in illegal activities increases significantly when multiple pieces of evidence are highly predicted, including the intention to invade prohibited areas; the intention to avoid surveillance areas; and the intention to conceal activities.

Below is a screenshot showing an example scenario from all conditions, along with the corresponding participant tasks.

\begin{figure}[H]
    \centering
    \includegraphics[width=0.7\textwidth]{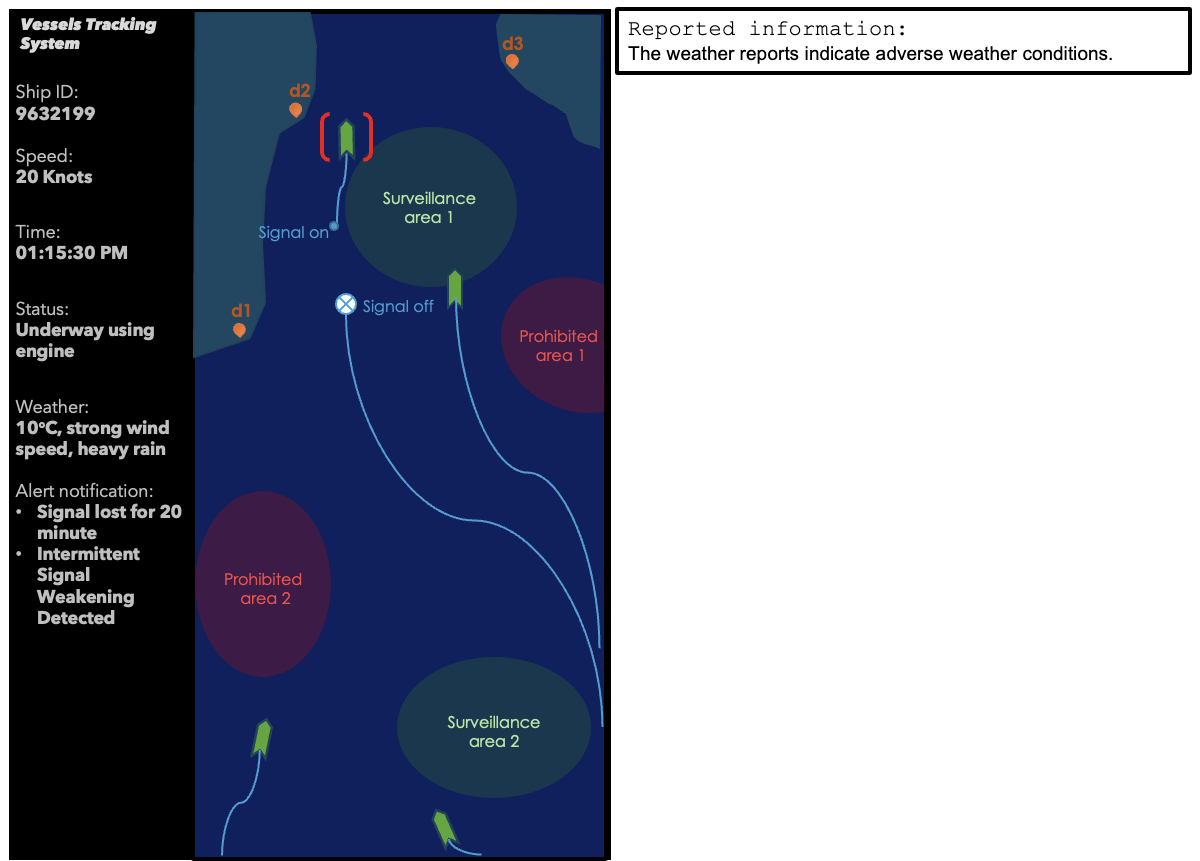} 
    \caption{Example scenario (NoGR Condition).}
    \label{fig:appendix_image7}
\end{figure}

\begin{figure}[H]
    \centering
    \includegraphics[width=0.7\textwidth]{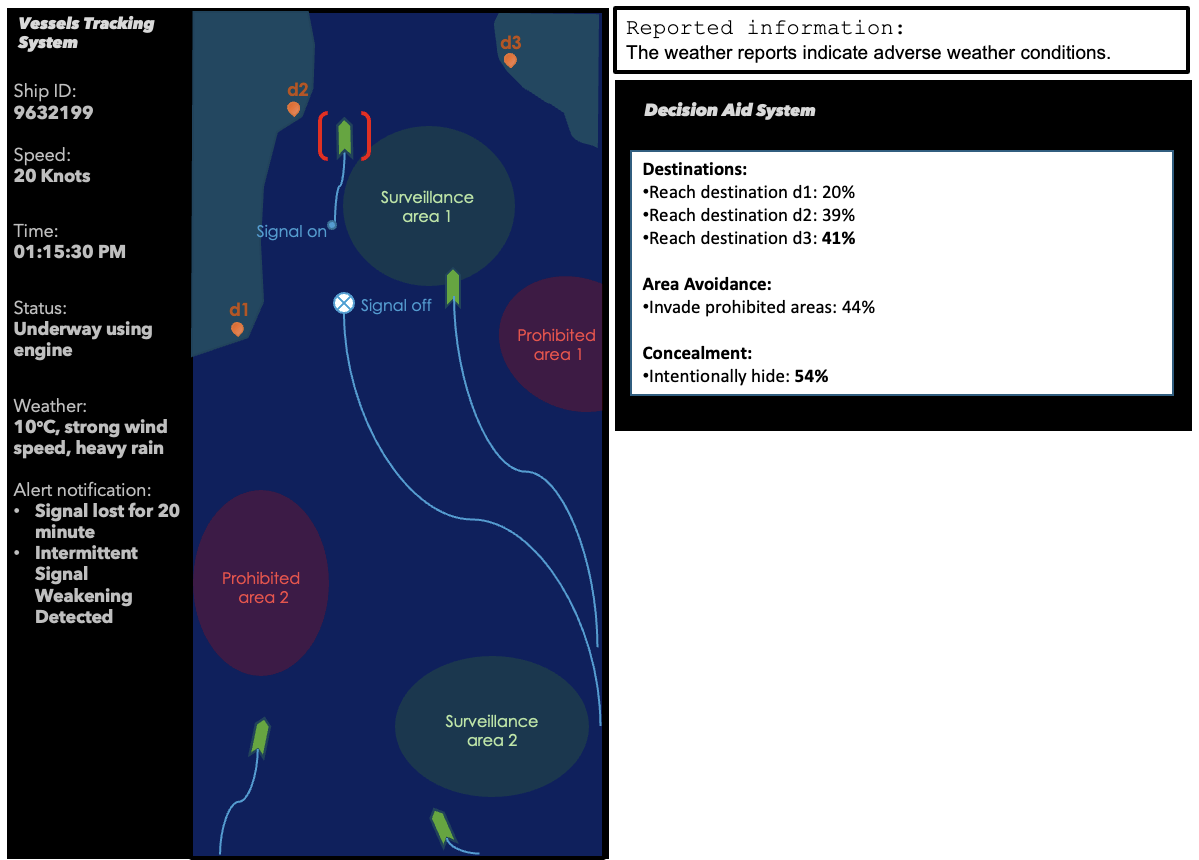} 
    \caption{Example scenario (GR Condition).}
    \label{fig:appendix_image8}
\end{figure}

\begin{figure}[H]
    \centering
    \includegraphics[width=0.7\textwidth]{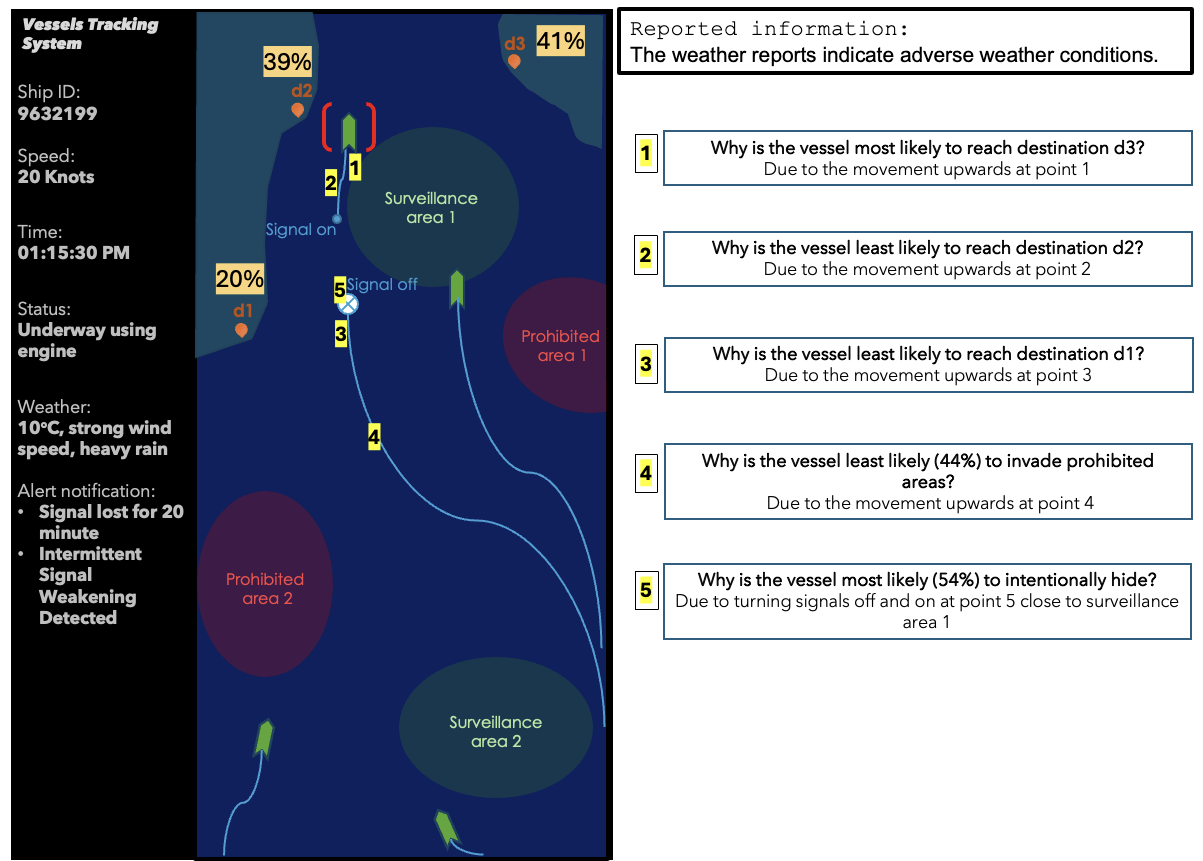} 
    \caption{Example scenario (XGR\_WoE Condition).}
    \label{fig:appendix_image9}
\end{figure}

\begin{figure}[H]
    \centering
    \includegraphics[width=0.7\textwidth]{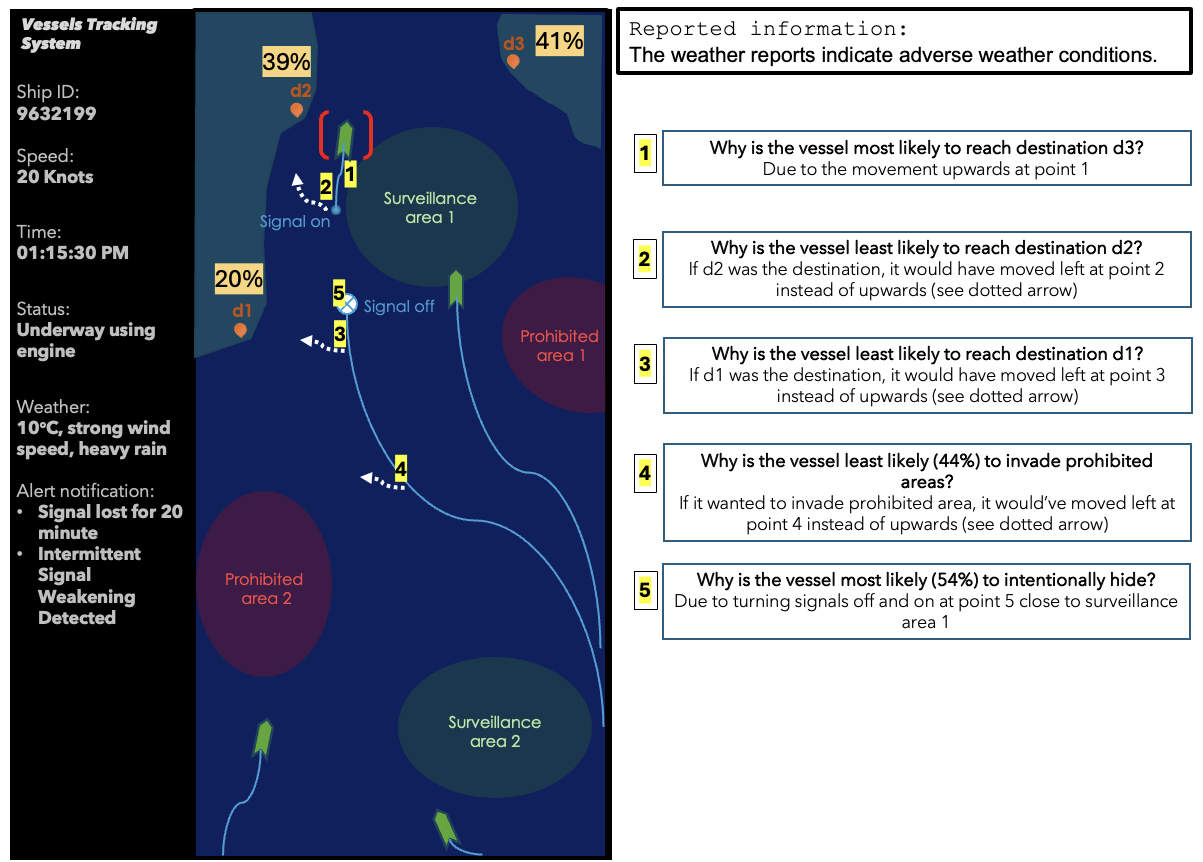} 
    \caption{Example scenario (XGR\_WoE\_CF Condition).}
    \label{fig:appendix_image10}
\end{figure}

\begin{figure}[H]
    \centering
    \includegraphics[width=0.8\textwidth]{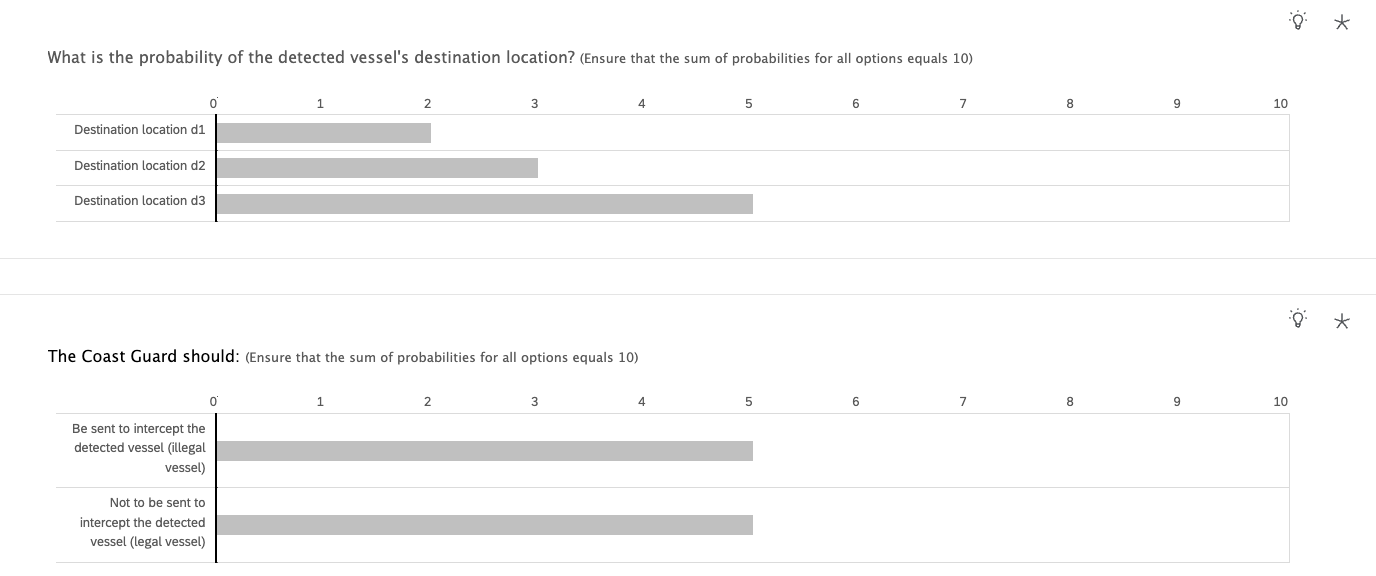} 
    \caption{Participant task, rate the likelihood of the vessel's destination and the necessity of intercepting the vessel on a scale from 0 to 10 (where 10 is most likely).}
    \label{fig:appendix_image11}
\end{figure}

\vskip 0.2in
\bibliography{sample}
\bibliographystyle{theapa}

\end{document}